\def\cvprPaperID{9317}
\def\confName{ICCV}
\def\confYear{2023}

\def\paperTitle{ReFit: Recurrent Fitting Network for 3D Human Recovery}

\def\authorBlock{
    Yufu Wang \qquad
    Kostas Daniilidis \\
    University of Pennsylvania \\
}

\newif\ifreview 
\newif\ifarxiv 
\newif\ifcamera \newcommand{\cameraready}{\cameratrue}
\newif\ifrebuttal 

\cameraready 

\pdfoutput=1
\documentclass[10pt,twocolumn,letterpaper]{article}
\ifreview \usepackage[review]{cvpr} \fi
\ifarxiv \usepackage[pagenumbers]{cvpr} \fi
\ifrebuttal \usepackage[rebuttal]{cvpr} \fi
\ifcamera \usepackage{cvpr} \fi

\usepackage{graphicx}
\usepackage{amsmath}
\usepackage{amssymb}
\usepackage{booktabs}

\usepackage{times}
\usepackage{microtype}
\usepackage{epsfig}
\usepackage[table,xcdraw]{xcolor}
\usepackage{caption}
\usepackage{float}
\usepackage{placeins}
\usepackage{color, colortbl}
\usepackage{stfloats}
\usepackage{enumitem}
\usepackage{tabularx}
\usepackage{xstring}
\usepackage{multirow}
\usepackage{xspace}
\usepackage{url}
\usepackage{subcaption}
\usepackage{xcolor}
\usepackage[hang,flushmargin]{footmisc}
\usepackage[accsupp]{axessibility}

\ifcamera \usepackage[accsupp]{axessibility} \fi

% ...

\ifarxiv  \fi

\newcommand{\R}[1]{{%
    \textbf{%
        \ifstrequal{#1}{1}{\textcolor{red}{R#1}}{%
        \ifstrequal{#1}{2}{\textcolor{blue}{R#1}}{%
        \ifstrequal{#1}{3}{\textcolor{magenta}{R#1}}{%
        \ifstrequal{#1}{4}{\textcolor{teal}{R#1}}{%
                           \textcolor{cyan}{R#1}%
        }}}}%
    }%
}}

\usepackage{xr-hyper}

\makeatletter
\newcommand*{\addFileDependency}[1]{
  \typeout{(#1)}
  \@addtofilelist{#1}
  \IfFileExists{#1}{}{\typeout{No file #1.}}
}

\makeatother

\usepackage[pagebackref,breaklinks,colorlinks]{hyperref}
\usepackage[capitalize]{cleveref}
\crefname{section}{Sec.}{Secs.}
\crefname{table}{Table}{Tables}
\crefname{figure}{Fig.}{Figs.}

\begin{document}
\title{\paperTitle}
\author{\authorBlock}
\maketitle

\begin{abstract}
We present Recurrent Fitting (ReFit), a neural network architecture for single-image, parametric 3D human reconstruction. ReFit learns a feedback-update loop that mirrors the strategy of solving an inverse problem through optimization. At each iterative step, it  reprojects keypoints from the human model to feature maps to query feedback, and uses a recurrent-based updater to adjust the model to fit the image better. Because ReFit encodes strong knowledge of the inverse problem, it is faster to train than previous regression models. At the same time, ReFit improves state-of-the-art performance on standard benchmarks. Moreover, ReFit applies to other optimization settings, such as multi-view fitting and single-view shape fitting. Project website: \href{https://yufu-wang.github.io/refit_humans/}{https://yufu-wang.github.io/refit\_humans/}


\end{abstract}
\section{Introduction}
\label{sec:intro}
Single-view 3D human reconstruction has transformed the way we analyze and create virtual content. Progress has primarily been driven by the use of parametric human models~\cite{loper2015smpl}, powerful neural networks~\cite{kanazawa2018end}, and high-quality annotated data~\cite{von2018recovering}. However, the best systems to date still struggle with various difficulties, including occlusion, uncommon poses, and diverse shape variations.

The traditional approach fits a parametric human model to an image using handcrafted objectives and energy minimization techniques~\cite{bogo2016keep}. But this optimization process is often riddled with local minima and corner cases. Because of such drawbacks, recent regression methods train a neural network to predict the parameters directly~\cite{kanazawa2018end, kolotouros2019learning}. Training such networks robustly requires a large amount of 3D annotated data that is hard to collect outside a lab setting. One line of research is to design better neural network architectures that can learn efficiently and generalize well to diverse in-the-wild cases. 

In this paper, we propose Recurrent Fitting (ReFit), an architecture for 3D human reconstruction. ReFit mimics the structure of model fitting, reducing the regression problem to a learning-to-optimize problem. This allows ReFit to learn faster than other architectures and improves state-of-the-art accuracy. 

\begin{figure}[]
    \center
    \vspace{1mm}
    \includegraphics[width=0.99\linewidth]{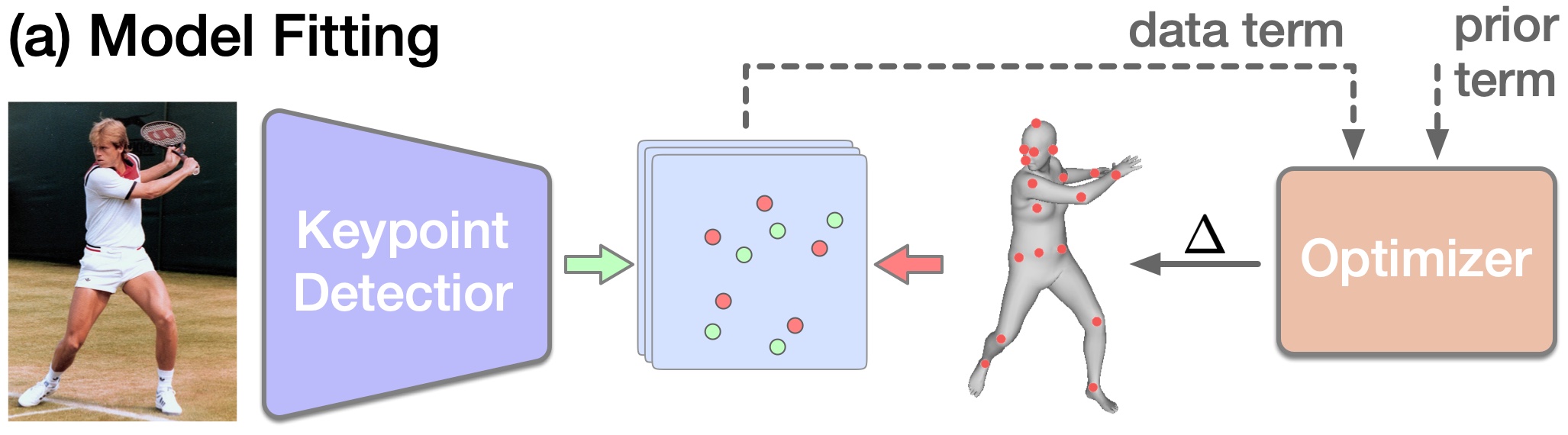} 
    \\
    \vspace{1.5mm}
    \includegraphics[width=0.99\linewidth]{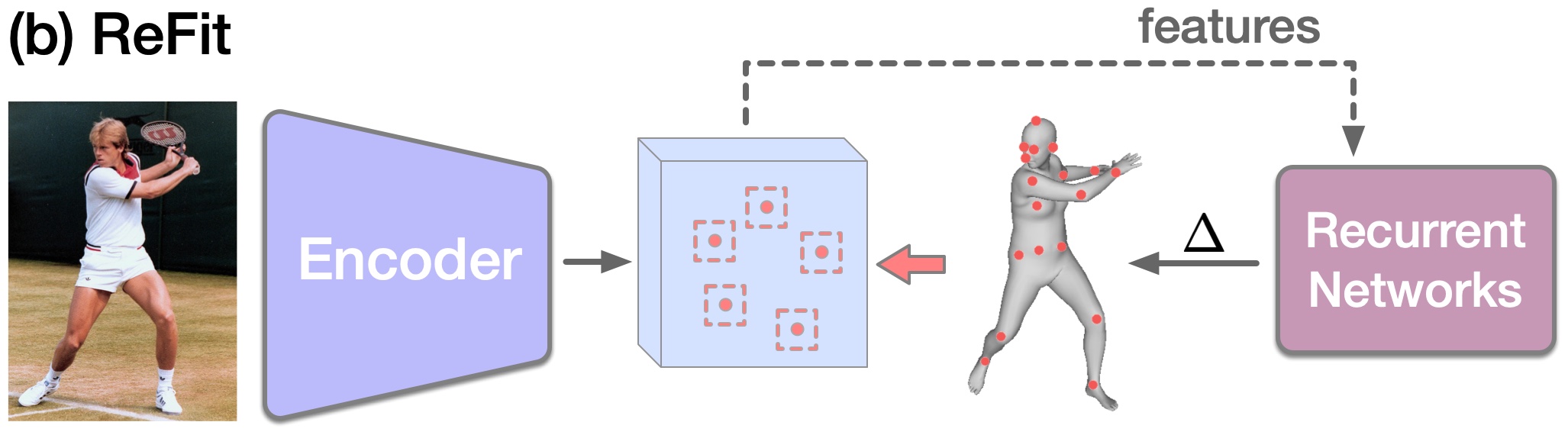}
    \vspace{-1mm}
\caption{\textbf{Overview}. ReFit mimics the strategy of model fitting by constructing a trainable feedback-update loop that adjusts the human model for a more accurate reconstruction. 
}
\label{fig:teaser}
\vspace{-2mm}
\end{figure}

ReFit has three main steps. (1) A backbone network extracts pixel-aligned image features. (2) A feedback module queries feedback features from keypoint reprojection. (3) A recurrent update module uses a set of disentangled GRUs to update the body. Feedback and updates are repeated until the body mesh is well-aligned with the image (Fig. \ref{fig:teaser}).

The design of ReFit mirrors traditional model fitting. In model fitting, the objective function computes the L2 distance between the reprojected and the detected keypoints~\cite{bogo2016keep}. In ReFit, each reprojected keypoint queries a window on a feature map. A window encodes first-order information, such as 2D flow, which is the derivative of the L2 objective. The queried features are fed to the update module to compute updates for the human model.

Another key element of ReFit is the disentangled and recurrent update streams. Iterative regression methods commonly map a feature vector to parameter updates with an MLP~\cite{kanazawa2018end, zhang2021pymaf}. Therefore, the update rules for the parameters are entangled until the last linear layer. In contrast, parameters in optimization follow update rules based on automatic differentiation, and the rules can be very different. As an example, in human pose, an alignment error in the right hand will produce a gradient for the right arm and shoulder but not for the left side. This notion of a disentangled update is difficult to achieve within a global feature vector. When learning the update function, we hypothesize that some degree of disentanglement is beneficial. 

To this end, ReFit's update module uses one GRU per parameter group to disentangle the parametric update. Each GRU learns the prior and the update rule for one parameter group, e.g., the rotation of a single joint. GRUs utilize memory from the previous iterations, analogous to a first-order method with momentum~\cite{choutas2022learning}. At each iteration, the GRUs can take different step sizes. In the example of right-hand misalignment, the GRUs for the right side of the body can output adjustments, while the GRUs for the left side can return little or no updates. This disentanglement leads to better performance. The disentangled GRUs can be implemented as batch matrix operations in PyTorch, allowing for efficient training and inference.  

Moreover, ReFit acts as a learned optimizer that applies to other settings of model fitting. Using ReFit as a learned optimizer, we propose a multi-view fitting procedure that significantly outperforms regression on separate views. We also demonstrate using ReFit to register a pre-scanned shape to images of the same subject, a helpful setting for outdoor motion capture.  

The ReFit architecture is substantially different from prior works. It uses a recurrent-based updater instead of a coarse-to-fine pyramid that limits the number of updates~\cite{zhang2021pymaf}. ReFit applies reprojection to query feedback instead of using a static global feature vector~\cite{zanfir2021neural}. Compared to learning-to-optimize approaches~\cite{song2020human, choutas2022learning}, ReFit does not detect keypoints but instead uses learned features and is end-to-end trainable.

ReFit improves state-of-the-art results on standard benchmarks~\cite{ionescu2013human3, von2018recovering}. The learning-to-optimize design accelerates training: ReFit converges with only 50K training iterations. We conduct extensive ablation studies to analyze the core components and applications, and contribute insights for future works.

\section{Related Work}
\label{sec:related}

\textbf{Human Mesh Model Fitting.} Reconstructing human pose and shape from images has a long history~\cite{hogg1983model,bregler1998tracking, blanz1999morphable, anguelov2005scape}. The reconstruction often is formulated as an energy minimization problem by fitting a parametric model with optimization in various settings: single view~\cite{sminchisescu2002human,bogo2016keep, pavlakos2019expressive,muller2021self, fan2021revitalizing}, multiple views~\cite{huang2017towards,dong2019fast,easymocap}, video~\cite{arnab2019exploiting, pavlakos2022human}, and other sensing inputs~\cite{von2018recovering, weiss2011home}. These optimization procedures involve multiple stages and sophisticated designs to avoid local minima.

The optimization objective commonly consists of a data term and a prior term. The data term measures the deviation between the estimation and the detected features, while the prior term imposes constraints on the pose and shape space. In practice, optimization encounters many difficulties, including noisy keypoint detection~\cite{cao2017realtime, sun2019deep}, complicated priors~\cite{pavlakos2019expressive, rempe2021humor, tiwari2022pose}, and the trade-off between the two terms. In the single-view setting where such difficulties compound, recent research shifts to two directions: learning to regress and learning to optimize. 

\textbf{Human Mesh Regression.} The recognition power of deep neural network~\cite{he2016deep}, paired with the representation capability of parametric human models~\cite{loper2015smpl, xu2020ghum}, has fueled the recent progress in single-view human mesh regression~\cite{kanazawa2018end, omran2018neural, kolotouros2019learning, kocabas2021pare, choi2020pose2mesh, lin2021end, jiang2020coherent, kanazawa2019learning, kocabas2020vibe, choi2021beyond, xiang2019monocular}. 

Iterative refinement is an essential strategy used by many regression works~\cite{kanazawa2018end, oberweger2015training, zanfir2021neural}. Carreira et al.~\cite{carreira2016human} motivate it from a neural science standpoint, which states that the human brain relies on feedback signals for various visual localization tasks. The de facto implementation introduced by HMR~\cite{kanazawa2018end} is to concatenate the prediction with the image feature vector to make new predictions recurrently. This alone is not an effective strategy, as the network has to learn error feedback solely in the latent space.

PyMAF~\cite{zhang2021pymaf, zhang2022pymaf} proposes a refinement strategy based on a coarse-to-fine feature pyramid. The parametric mesh predicted from the coarse level is reprojected to feature maps at the finer levels to gather spatial features for refinement updates. This strategy lifts the error feedback to the image space, but the pyramid limits the number of update steps.

ReFit reprojects mesh keypoints to the feature maps at a single resolution but constructs a recurrent loop with GRUs that is not limited to the number of pyramid levels. Moreover, ReFit's feedback and update modules have novel designs that further boost the accuracy. 

\textbf{Learning to Optimize.} Many vision problems are inverse problems traditionally solved by optimization. Learning to optimize aims to train a neural network to propose a descend direction that replaces or supplements the optimizer update~\cite{adler2017solving}. This paradigm inspires network designs that mimic  optimizers in various vision tasks~\cite{adler2018learned, lv2019taking, flynn2019deepview, teed2020raft, lipson2022coupled}. 

For the human model, LGD~\cite{song2020human} and LFMM~\cite{choutas2022learning} improve traditional model fitting by using a neural network to predict residues for the optimizer updates. This approach is not end-to-end differentiable because the optimizer uses keypoint detection to compute the gradients. Neural Descent~\cite{zanfir2021neural} replaces the optimizer with an LSTM-based updater~\cite{hochreiter1997long} and reduces the detections into a feature vector. However, reducing spatial representation into a feature vector prevents it from having direct error feedback in the image space. ReFit is an evolution of these methods. It uses gated recurrent units~\cite{cho2014properties, chung2014empirical} as updaters but utilizes reprojection on the learned feature maps as the feedback to infer an update.

ReFit can be regarded as a learned optimizer. Beside single-view inference, we demonstrate that ReFit can be a drop-in replacement for traditional optimizers in multi-view model fitting. We also apply ReFit to register a pre-fitted shape to images of the same subject.


\section{Recurrent Fitting}
\label{sec:method}
\begin{figure*}[!t]
	\centering
    \includegraphics[width=0.96\textwidth]{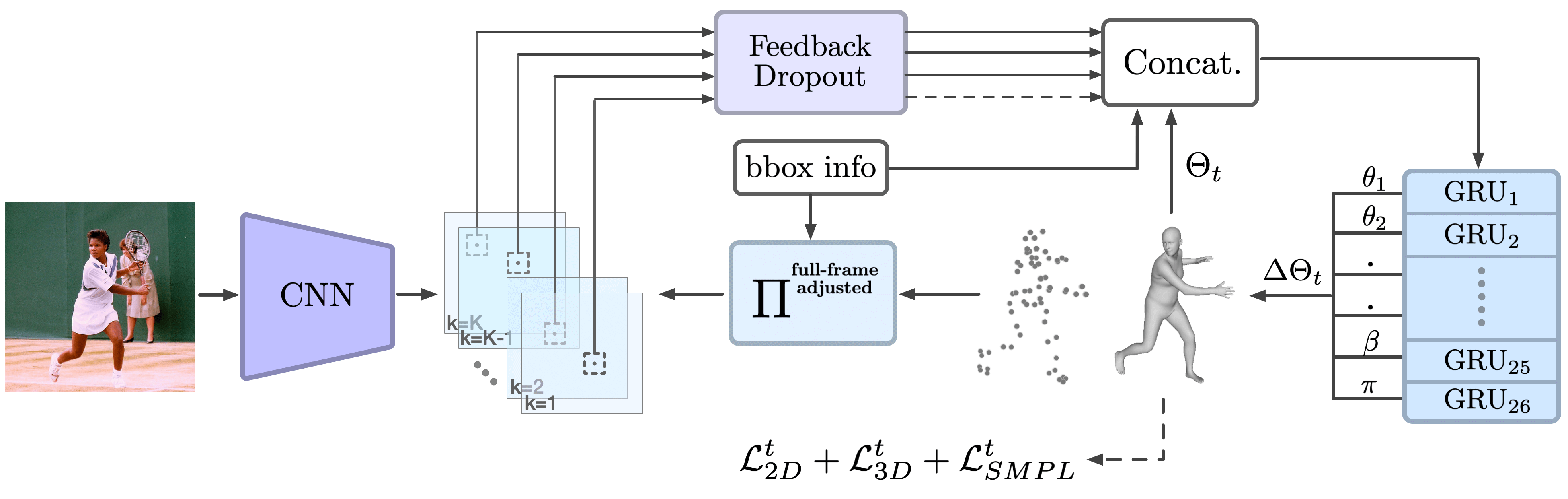}
	\vspace{-2mm}
	\caption{\textbf{The ReFit Network.} ReFit extracts one feature map per keypoint with a backbone network (Sec.~\ref{subsec:feature}). It then reprojects keypoints from the 3D human mesh to the corresponding feature maps using the full-frame adjusted camera model (Sec.~\ref{subsec:feedback}). Feedback is dropped randomly during training, and concatenated with the current estimate $\Theta_t$ and the bounding box info to form the final feature vector. The final feature is sent to $N$ parallel GRUs to predict updates for the $N$ parameters (Sec.~\ref{subsec:update}). The updated mesh is again reprojected to the feature maps to repeat the feedback-update loop until good reconstruction is achieved.}
\label{fig:method}
\vspace{-2mm}
\end{figure*}

Given an image of a person, our goal is to predict the parameters of the SMPL human model. The ReFit network extracts image features (Sec. \ref{subsec:feature}), compares the human mesh with the pixel-aligned features (Sec. \ref{subsec:feedback}), and iteratively updates the SMPL parameters (Sec. \ref{subsec:update}).  An overview of the method is given in Figure~\ref{fig:method}. \smallbreak

\textbf{Preliminaries}. The SMPL model~\cite{loper2015smpl} is parametrized with $\Theta = \{ \theta, \beta, \pi \}$, where $\theta \in \mathbb{R}^{24\times3}$ are the relative rotations of the 24 body joints, $\beta \in \mathbb{R}^{10}$ is the shape parameter, and $\pi \in \mathbb{R}^{3}$ is the root translation w.r.t the camera. Given $\Theta$, SMPL outputs the 3D mesh $\mathcal{M}(\Theta) \in \mathbb{R}^{6890\times3}$. 

\subsection{Feature Extraction}
\label{subsec:feature}
We use High-Resolution Net (HRNet)~\cite{sun2019deep} as the feature extractor. It produces spatially precise feature maps that are beneficial for the feedback step. Given an image $I \in \mathbb{R}^{H\times W\times3}$, it outputs feature maps $F\in \mathbb{R}^{H/4\times W/4\times K}$, where $K$ is the number of channels. We set $K$ to equal the number of keypoints. 

In addition, we average pool the low-resolution branch of HRNet to produce a global feature vector and add a linear layer to predict an initialization $\Theta_0$ for the parameters.

\subsection{Feedback}
\label{subsec:feedback}
Given an initial estimate, we reproject keypoints from the SMPL model to the feature maps $F$ to retrieve spatial features. Each keypoint is only reprojected to one channel, yielding $K$ channels for $K$ keypoints. This design is motivated by model fitting, where the keypoint detector outputs one channel per keypoint. 

For a reprojection $x_k = (u, v)$ where $u$ and $v$ are the pixel coordinates, we take the feature values inside a window centered at $x_k$ as
\begin{equation}
f_k = \{f(x) \in F_k |  \parallel x-x_k \parallel \leq r \}
\label{eq:window}
\end{equation}
where we set $r=3$ pixels as the radius of the window. We concatenate feedback from all the keypoints, along with the current estimate $\Theta_t$ and the bounding box center $c^{bbox}$ and scale $s^{bbox}$, to form the final feedback vector as
\begin{equation}
f = [f_1, ..., f_K, \Theta_t, c^{bbox}, s^{bbox}]
\label{eq:feedback}
\end{equation}

\begin{figure}[]
    \center
	\vspace{-2mm}
    \includegraphics[width=0.15\linewidth]{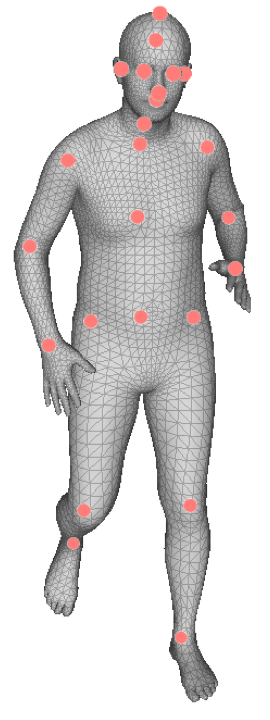} 
    \hspace{8mm}
    \includegraphics[width=0.15\linewidth]{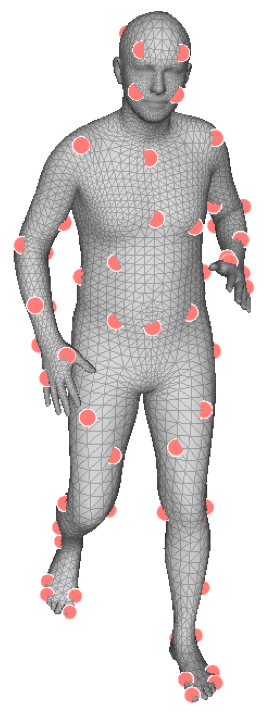}
    \hspace{8mm}
    \includegraphics[width=0.15\linewidth]{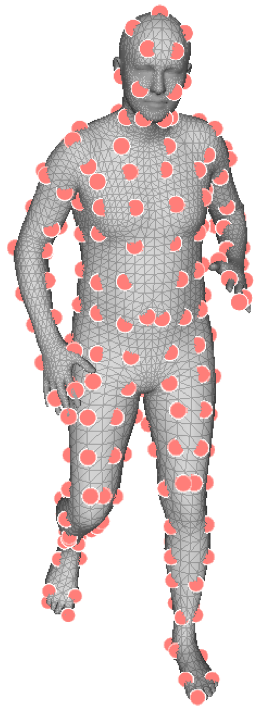}
    \vspace{-3mm}
\caption{\textbf{Types of keypoints.} From left to right are semantic keypoints, mocap markers, and uniformly sampled vertices. We use one of the three types during feedback (Sec.~\ref{subsec:feedback}).
}
\label{fig:keypoint}
\vspace{-2mm}
\end{figure}

\textbf{Types of Keypoints}. The per-keypoint feature map does not directly detect a keypoint. Instead, each channel learns the features associated with a keypoint. This process does not require direct supervision, which allows us to test different types of keypoints where 2D annotations are unavailable. We test three types (Figure~\ref{fig:keypoint}): semantic keypoints ($K$ = 24); mocap markers ($K$ = 67); and evenly sampled mesh vertices ($K$ = 231). We examine the three types separately and do not combine them.

Of the three, mocap markers provide better pose and shape information than semantic keypoints, and are less redundant than mesh vertices. We use the same mocap markers as in AMASS~\cite{mahmood2019amass}, and the markers are defined on the mesh surface by selecting the closest vertices.

\textbf{Full-frame Adjusted Reprojection}. Human regression methods take a square-cropped image of a person as input, assuming the optical axis going through the crop center. But the input is usually cropped from a full-frame image with a different optical axis. This deviation incurs an error in the global rotation estimation, and by implication, the body pose estimation.

CLIFF~\cite{li2022cliff} proposes to supervise the 2D reprojection loss in the original full-frame image, by using the full-frame reprojection:
\begin{equation}
x^{full}_{2D} = \Pi(X^{full}_{3D}) = \Pi(X_{3D} + t^{full})
\label{eq:full-frame}
\end{equation}
where $\Pi$ is the perspective projection using the original camera intrinsics, $X_{3D}$ are the body keypoints in the canonical body coordinate, and $t^{full}$ is the translation with respect to the optical center of the original image. The camera intrinsics, if unknown, can be estimated from the dimensions of the full image~\cite{kissos2020beyond}. This reprojection is more faithful to the image formation process and leads to a better global rotation estimation.

We propose to adjust the points back to the cropped image after full-frame reprojection:
\begin{equation}
x^{crop}_{2D} = (x^{full}_{2D} - c^{bbox}) / s^{bbox}
\label{eq:full-frame}
\end{equation}
where $c^{bbox}$ and $s^{bbox}$ are the location and the size of the bounding box from which the crop is obtained. We call this \emph{full-frame adjusted reprojection}. This seemingly trivial operation grants two advantages. First, the scale of the reprojection is normalized. The reprojection error is unaffected by the size of the person in the original image. 

But more importantly, it extends full-frame reprojection to the feedback step. We reproject keypoints to the cropped image feature maps, but with this camera model, the locations are properly adjusted to be consistent with full-frame reprojection. We use this model to retrieve $f_k$ for each keypoint as in Eq \ref{eq:window}.

\textbf{Feedback Dropout}. Each keypoint reprojection produces a feedback signal. To combine signals from all the keypoints, we formulate it as ensemble learning using dropout~\cite{srivastava2014dropout}. Specifically, we add a dropout layer, where there is a $p$ = 0.25 chance that feedback $f_k$ will be zeroed out during training.

During training, the network learns to infer the pose and shape using subsets of keypoints. This prevents co-adaptations~\cite{hinton2012improving} of feedback signals, and makes the network robust at test time when some keypoints are occluded. 

This design also has ties to Spatial Dropout~\cite{tompson2015efficient}, which randomly drops feature map channels as opposed to pixel-wise activations. Similarly, we drop a keypoint's feedback completely instead of dropping some values.

\subsection{Update}
\label{subsec:update}
The update module takes the feedback signal $f$ as input (Eq.~\ref{eq:feedback}), and predicts an update step $\Delta\Theta_t$, which is added to produce the next estimate: $\Theta_t + \Delta\Theta_t \rightarrow \Theta_{t+1}$.

The update is split into $N=26$ parallel streams. Each stream is responsible for updating one SMPL parameter. There are 26 streams: 24 for the joint rotations, one for the shape parameter, and one for the translation.

Each stream has an identical structure, consisting of a GRU and a 2-layer MLP. The GRU updates its hidden state: $(f, h^n_{t-1}) \rightarrow h^n_t$, and the MLP maps the hidden state to the parametric update: $(h^n_t) \rightarrow \Delta\theta^n_t$. The parallel streams can be implemented as batch matrix operations to run efficiently. 

The update module acts as a first-order optimizer. Each stream learns the update rule of a parameter. Multiple streams effectively disentangle the update rules. The feedback signal is similar to the data term that informs the error. The prior, which is hand-crafted in optimization, is now learned by the update module. 

Having 26 update streams does not increase the complexity because we also reduce the hidden units. The complexity per layer is $O(NM^2)$, where $N$ is the number of streams and $M$ is the size of the hidden layer. Previous works use $N=1$ and $M=1024$, leading to 1M parameters per layer. We use $N=26$ and $M=32$, which are 27K parameters per layer.

\subsection{Supervision}
\label{subsec:supervision}
The iterative update steps produce a sequence of estimates $\{\Theta_0, ..., \Theta_T\}$. At inference time, $\Theta_T$ is the final prediction. During training, we supervise all the iterations. 

At each iteration $t$, the loss is made up of three terms,
\begin{equation}
\mathcal{L}_t = \lambda_{2D}\mathcal{L}^t_{2D} + \lambda_{3D}\mathcal{L}^t_{3D} + \lambda_{SMPL}\mathcal{L}^t_{SMPL}
\label{eq:loss}
\end{equation}
where each term is calculated as 
\begin{align*}
\mathcal{L}^t_{2D} &= ||\mathcal{\hat{J}}_{2D} - \Pi(\mathcal{J}^t_{3D})||^2_F \\
\mathcal{L}^t_{3D} &= ||\mathcal{\hat{J}}_{3D} - \mathcal{J}^t_{3D}||^2_F \\
\mathcal{L}^t_{SMPL} &= ||\hat{\Theta} - \Theta_t||^2_2
\label{eq:terms}
\end{align*}
$\mathcal{J}_{3D}$ are the 3D joints obtained from the SMPL model, the hat operator denotes the ground truth of that variable, and $\Pi$ is the full-frame adjusted reprojection. 

The final loss is a weighted sum of the loss at each iterative update
\begin{equation}
\mathcal{L} = \sum_{t=0}^{T}\gamma^{T-t}\mathcal{L}_t 
\label{eq:loss}
\end{equation}
where we set $\gamma=0.85$ and $T=5$ for all experiments. Prior works supervise only the last iteration to prevent the overshoot behavior, but it requires the gradient to flow through a long sequence of estimates, which slows down training convergence. 

Our supervision is inspired by RAFT~\cite{teed2020raft}, an iterative optical flow method. First, the gradient is backpropagated through $\Delta\Theta_t$ but not through $\Theta_t$. In other words, at each iteration, we only supervise the update but not the prediction from the previous step. Second, we use the proposed weighted sum that downscales the importance of earlier iterations.

\subsection{Applications}
\label{subsec:applications}
ReFit is trained with single-view images for single-view inference. But since ReFit operates similarly to an optimizer, we demonstrate two applications in which traditionally an optimizer is used: multi-view model fitting, and fitting a pre-acquired shape to images of the same subject.

\begin{figure}[]
    \center
    \includegraphics[width=0.90\linewidth]{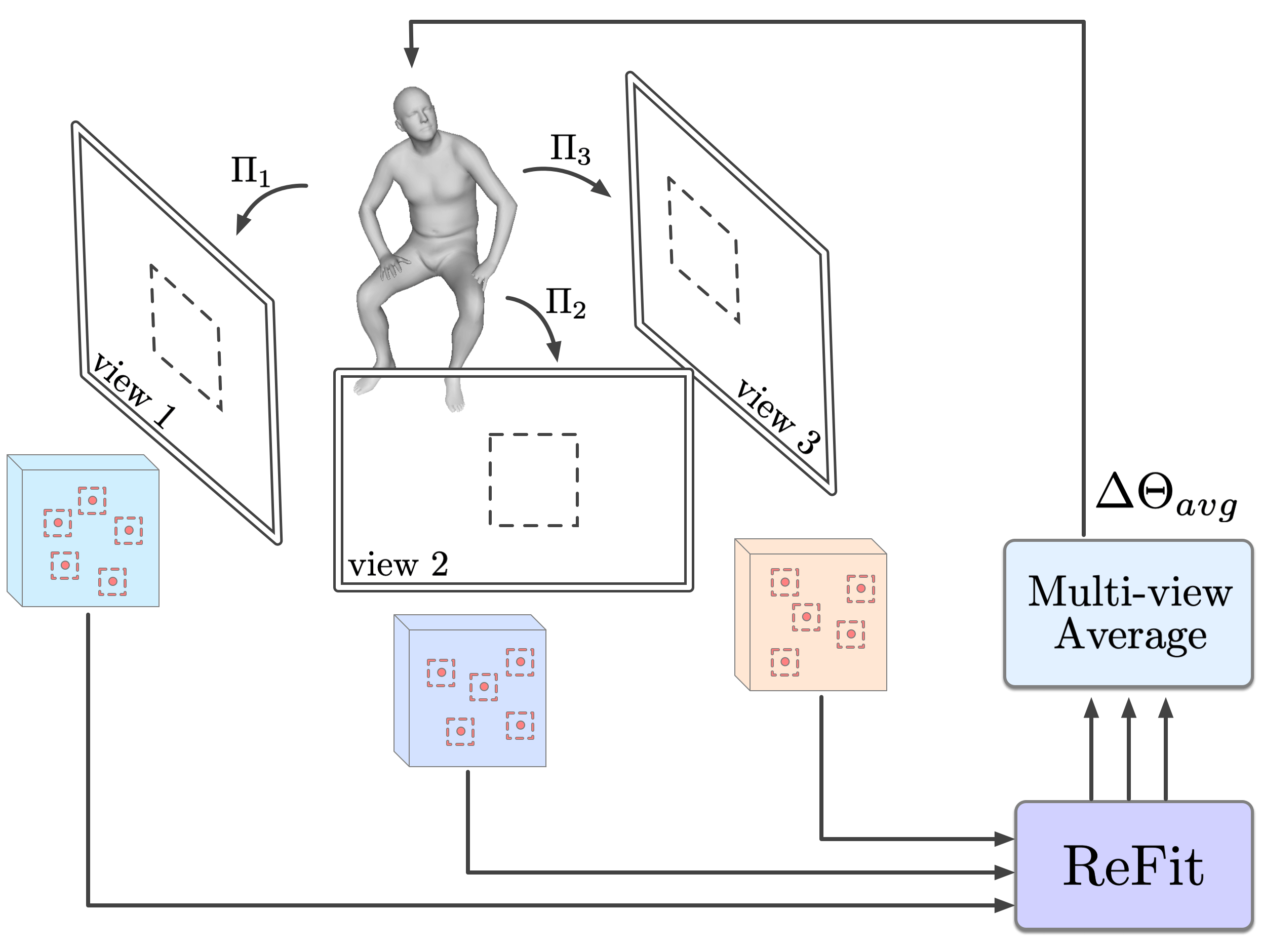} 
    \vspace{-2mm}
\caption{\textbf{Multi-view ReFit.} At each iteration, ReFit operates on all views independently to produce updates. The multi-view averaging procedure pool updates across views to produce a single update for the mesh. The updated mesh is again reprojected to each view to repeat the iterative process (Sec.~\ref{subsec:applications}). 
}
\label{fig:method_multiview}
\vspace{-2mm}
\end{figure}

\textbf{Multi-view ReFit}. Motion capture with multiple calibrated cameras offers the highest accuracy~\cite{ionescu2013human3, dong2019fast}. In a markerless setting, an optimization procedure detects keypoints in each view and fits the model by minimizing the reprojection error~\cite{easymocap}. We replace this procedure using ReFit as shown in Figure~\ref{fig:method_multiview}.

ReFit operates on each view independently to produce updates. The updates are averaged across views with a multi-view averaging procedure. We average the shapes by taking the mean of the predicted PCA coefficients across views. Because the body pose consists of the relative rotation of each joint, we directly average the body poses across views with rotation averaging, which averages the rotation matrices and uses SVD to project the result back to SO(3). The predicted global rotations are with respect to each view, so we transform them to a global coordinate with the calibrated extrinsic before performing rotation averaging. This procedure results in a single copy of the updated model. The model is then reprojected to all views to repeat the feedback-update step.

The difference between multi-view ReFit versus simply averaging the final predictions is that the averaging happens during fitting. So at each iteration, the update is in the direction that explains multiple views.

\textbf{Shape ReFit}. In this setting, we aim to fit a known body shape to images of the same subject with different poses. This is relevant for applications where the shape of a subject can be pre-scanned or pre-fitted and then used for outdoor motion capture (mocap)~\cite{von2018recovering, habermann2020deepcap}. Registering the shape to its images is commonly done with optimization, but we demonstrate that ReFit can perform a similar function.

We call this procedure Shape ReFit. At each iteration, we ignore the shape prediction and use the known shape to render the SMPL model. Therefore, the reprojection reflects the alignment status of the ground truth shape with the image, and ReFit will adjust the pose and position of the model for better alignment. 

We perform experiments to verify that Shape ReFit can indeed fit a known and fixed shape to its images.

\section{Experiments}
\label{sec:experiments}
\textbf{Datasets.} We train the ReFit model with 3DPW~\cite{von2018recovering}, Human3.6M~\cite{ionescu2013human3}, MPI-INF-3DHP~\cite{mehta2017monocular}, COCO~\cite{lin2014microsoft} and MPII~\cite{andriluka20142d}. We use the pseudo ground truth SMPL annotations from EFT~\cite{joo2021exemplar} for COCO and MPII. We evaluate ReFit on 3DPW and Human3.6M, with the MPJPE (mean per-joint error) and PA-MPJPE (Procrustes-aligned) metrics. \smallbreak

\textbf{Implementation.} We train ReFit for 50k iterations with all the datasets and evaluate on 3DPW. We then finetune the model with only Human3.6M for another 50k iterations and evaluate on Human3.6M. We use the Adam optimizer with a learning rate of 1e-4 and a batch size of 64. The input image is resized to 256$\times$256. Standard data augmentations are applied. We include additional details in the supplementary. 

\subsection{Quantitative Evaluation}
\label{subsec:quantitative}
We present quantitative evaluation in Table \ref{tab:evaluation}. Our method outperforms all single-view methods. 

We organize the methods by their training data to present a moderate view of progress: the quality of training data plays a major role in improving test accuracy, as is pointed out by recent studies~\cite{pang2022benchmarking}. We follow this practice and train CLIFF~\cite{li2022cliff} as our baseline with the same data and procedure. Overall, ReFit outperforms other architectures. 

We exam the generalization by training with synthetic data from BEDLAM~\cite{black2023bedlam} following their proposed losses. Table \ref{tab:bedlam} shows that ReFit generalizes better, and by using both synthetic and real data for training, it achieves the best results.

\begin{table}[ht!]
\centering
\footnotesize
\hspace{-3mm}
\tabcolsep=0.85mm

\begin{tabular}{p{0.18\linewidth} p{0.20\linewidth} cccc@{}} 
\toprule
 \multirow{2}{*}{\small Training} & \multirow{2}{*}{\small Architecture} & \multicolumn{2}{c}{3DPW} & \multicolumn{2}{c}{Human3.6M} \\
\cmidrule(lr){3-4} \cmidrule(lr){5-6} 
 & & MPJPE & PA-MPJPE & MPJPE & PA-MPJPE \\ 
\midrule
 


H+M+2D &  HMR~\cite{kanazawa2018end} & 130.0 &  76.7 & 88.0 & 56.8 \\ 
    
\midrule
 \multirow{3}{*}{\shortstack{H+M+\\2D-SPIN~\cite{kolotouros2019learning}}} 
    & HMR$^\dagger$ & 98.5 & 60.9 & 64.8 & 43.7 \\
    & ProHMR~\cite{kolotouros2021probabilistic} & - & 59.8 & - & 41.2 \\
    & PyMAF~\cite{zhang2021pymaf} & 92.8 & 58.9 & 57.7 & 40.5 \\

\midrule
 \multirow{4}{*}{\shortstack{H+M+\\2D-EFT~\cite{joo2021exemplar}}} 
    & PARE$^*$~\cite{kocabas2021pare}  & 82.0 & 50.9 & - & - \\
    & HMR$^{*\dagger}$   & 79.5 & 48.0 & 58.8 & 39.5 \\
    & PyMAF$^*$~\cite{zhang2022pymaf} & 78.0 & 47.1 & 54.2 & 37.2 \\
    & \textbf{ReFit$^*$}  & \textbf{71.0} & \textbf{43.9} & \textbf{48.5} & \textbf{32.4} \\
    
\midrule
 \multirow{5}{*}{\shortstack{H+M+\\3DPW+\\2D-EFT~\cite{joo2021exemplar}}} 
    & PARE$^*$~\cite{kocabas2021pare} & 74.5 & 46.5 & - & - \\
    & PyMAF$^*$~\cite{zhang2022pymaf} & 74.2 & 45.3 & - & - \\
    & HybrIK~\cite{li2021hybrik} & 74.1 & 45.0 & 55.4 & 33.6 \\
    & CLIFF$^{*\ddagger}$~\cite{li2022cliff} & 73.5 & 44.3 & 52.6 & 35.0 \\
    & \textbf{ReFit$^*$} & \textbf{65.8} & \textbf{41.0} & \textbf{48.4} & \textbf{32.2} \\

\bottomrule
\end{tabular}
\vspace{-2mm}
\caption{\small \textbf{Evaluation} grouped by training data (H: H36M, M: MPI-INF-3DHP, 2D-X: 2D datasets with 3D pseudo-gt from method X). The superscripts denote ($*$: using HRNet backbone, $\dagger$: implementation from Zhang et al.~\cite{zhang2021pymaf, zhang2022pymaf}, $\ddagger$: our implementation).}
\vspace{-3mm}
\label{tab:evaluation}
\end{table}


\begin{table}[ht!]
\centering
\footnotesize
\vspace{2mm}
\tabcolsep=0.85mm
\begin{tabular}{p{0.25\linewidth} p{0.30\linewidth} p{0.10\linewidth} c p{0.08\linewidth}@{}} 
\toprule
 \multirow{2}{*}{\small Training} & \multirow{2}{*}{\small Architecture} & \multicolumn{3}{c}{3DPW}\\
 \cmidrule(lr){3-5} & & MPJPE & PA-MPJPE & PVE \\ 
\midrule
 \multirow{2}{*}{BEDLAM~\cite{black2023bedlam}} 
    & BEDLAM-CLIFF & 72.0 & 46.6 & 85.0  \\
    & \textbf{ReFit} & \textbf{66.2} & \textbf{43.8} & \textbf{80.1}  \\
\midrule
 BEDLAM + Real & \textbf{ReFit} & \textbf{57.6} & \textbf{38.2} & \textbf{67.6} \\ 
\bottomrule
\end{tabular}
\vspace{-2mm}
\caption{\small \textbf{Evaluation}: with additional synthetic data from BEDLAM.}
\vspace{-3mm}
\label{tab:bedlam}
\end{table}


\subsection{Ablations}
\label{subsec:ablations}
We conduct ablation experiments to examine the core components of ReFit. All models are trained with the same data as the proposed main model, and tested on 3DPW. 

\textbf{Type of Keypoints.} We train three models with the three types of keypoints in the feedback step. They are used for reprojection to query features, but there is no supervision on the 2D locations. Using mocap markers produces the best results, confirming previous studies~\cite{mahmood2019amass, zhang2021we} that demonstrate mocap markers as a good proxy to infer the human body. \smallbreak 

\textbf{Feedback Dropout.} We test the effect of dropout in the feedback step. We see that feedback dropout significantly boosts accuracy, likely because it prevents co-adaptations of keypoint signals, and makes the model robust when keypoints are occluded. \smallbreak

\textbf{Full-frame Adjusted Reprojection.} Our result confirms that using full-frame reprojection for supervision improves accuracy~\cite{li2022cliff}. The proposed full-frame adjusted model extends to the feedback step. The best result is achieved when the camera model is faithful to the full image formation in all stages of the network. \smallbreak

\begin{table}[ht!]
\centering
\footnotesize
\hspace{-3mm}
\tabcolsep=0.85mm

\begin{tabular}{@{}>{\centering\arraybackslash}p{0.32\linewidth} p{0.30\linewidth} >{\centering\arraybackslash}p{0.15\linewidth} >{\centering\arraybackslash}p{0.20\linewidth}@{}} 
\toprule
 \multirow{2}{*}{\small Experiment} & \multirow{2}{*}{\small Method} & \multicolumn{2}{c}{3DPW}\\
\cmidrule(lr){3-4}
 & & MPJPE & PA-MPJPE \\ 
\midrule
 
 \multirow{3}{*}{\shortstack{Type of Keypoints\\Reprojection}}
    & Semantic Keypoints & 70.3 & 42.1 \\ 
    & \textbf{Mocap Markers} & 65.8 & 41.0 \\
    & Sparse Vertices & 69.2 & 41.9 \\
    
\midrule
 \multirow{3}{*}{Feedback Dropout} 
    & No Dropout & 70.2 & 42.1  \\
    & $p = 0.15$ & 68.5 & 41.5  \\
    & $\boldsymbol{p = 0.25}$ & 65.8 & 41.0  \\

\midrule
 \multirow{3}{*}{\shortstack{Full-frame Adjusted\\Reprojection} }
    & No Full-frame & 70.6 & 42.6  \\
    & Only Supervision & 68.8 & 42.5  \\
    & \textbf{Supervision+Feedback} & 65.8 & 41.0  \\

\midrule
 \multirow{3}{*}{Feedback Radius} 
    & $r = 0$ & 68.7 & 42.4  \\
    & $r = 1$ & 68.6 & 41.7  \\
    & $\boldsymbol{r = 3}$ & 65.8 & 41.0  \\

\midrule
 \multirow{4}{*}{\shortstack{Inference\\Iterations} } 
    & $T = 0$ & 73.0 & 45.0  \\
    & $T = 2$ & 67.6 & 42.1  \\
    & $\boldsymbol{T = 5}$ & 65.8 & 41.0  \\
    & $T = 10$ & 66.9 & 42.0  \\

\midrule
 \multirow{2}{*}{Update Module} 
    & One GRU & 69.6 & 42.3  \\
    & \textbf{26 GRUs} & 65.8 & 41.0  \\

\midrule
 \multirow{2}{*}{Supervision} 
    & Last Iteration & 70.5 & 44.3  \\
    & \textbf{All Iterations} & 65.8 & 41.0  \\

\bottomrule
\end{tabular}

\vspace{-2mm}
\caption{\small \textbf{Ablation of model designs}. The highlighted option is used for the final model. We detail each experiment in Sec. \ref{subsec:ablations}.
}
\label{tab:ablation}
\end{table}


\textbf{Feedback Radius.} For each keypoint, we query a window at its reprojection location as the feedback feature. The radius of the window affects the local context. The motivation is that the window encodes first-order information to indicate where the keypoint should move on the 2D image plane. Overall, we find that $r=3$ works well. \smallbreak

\textbf{Inference Iterations.} We train the model with $T=5$ update steps during training. At inference time, we test the model accuracy with various update steps. $T=0$ indicates the initial guess from the backbone. Most previous methods use $3$ steps, corresponding to $T=2$ in our case. We see the benefit of using more update steps.  We also observe that increasing to $T=10$ does not cause the estimation to diverge. \smallbreak

\textbf{Update Module.} We use 26 parallel GRU streams to predict updates. To test the alternative, we swap out the 26 GRU with one larger GRU with 516 hidden units. This model has lower accuracy, confirming our hypothesis that it is beneficial to have separate update operators to learn separate update rules. 

\textbf{Supervision.} We test supervising all iterations against supervising only the last iteration. Supervising all iterations achieves higher accuracy, but stopping the gradient across iterations as stated in Sec. \ref{subsec:supervision} is important for stable training. 

\subsection{Qualitative Evaluation}
\label{subsec:qualitative}
\begin{figure*}[!t]
\vspace{-1mm}
\captionsetup[subfigure]{labelformat=empty}
    \centering
    \begin{subfigure}[t]{0.245\textwidth}
    \caption{ \textbf{30$^{\text{th}}$ percentile }}
            \includegraphics[width=0.49\textwidth]{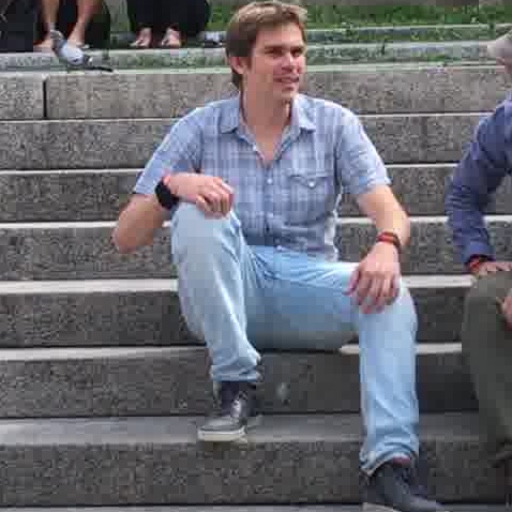}
            \includegraphics[width=0.49\textwidth]{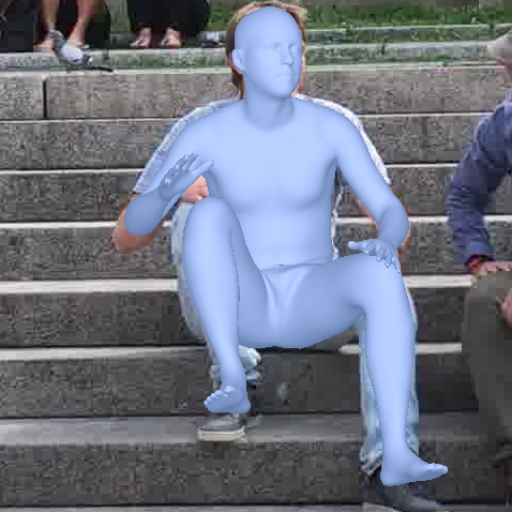}
   \end{subfigure}
    \begin{subfigure}[t]{0.245\textwidth}
    \caption{ \textbf{60$^{\text{th}}$ percentile }}
            \includegraphics[width=0.49\textwidth]{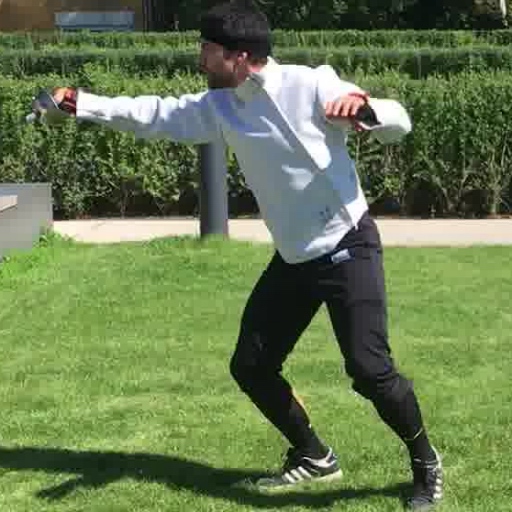}
            \includegraphics[width=0.49\textwidth]{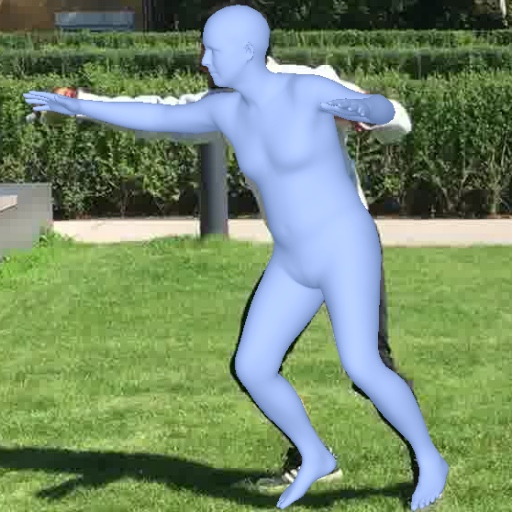}
   \end{subfigure}
    \begin{subfigure}[t]{0.245\textwidth}
    \caption{ \textbf{90$^{\text{th}}$ percentile }}
            \includegraphics[width=0.49\textwidth]{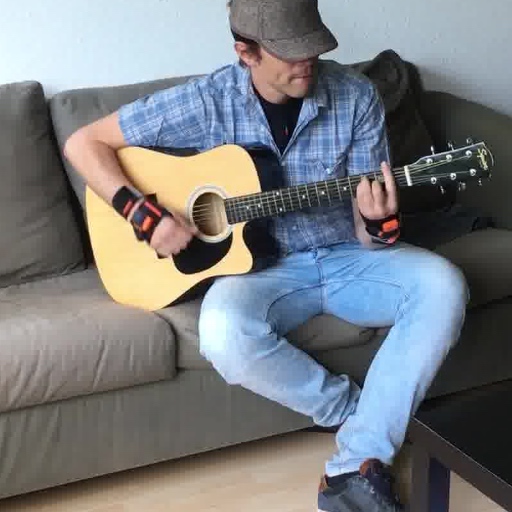}
            \includegraphics[width=0.49\textwidth]{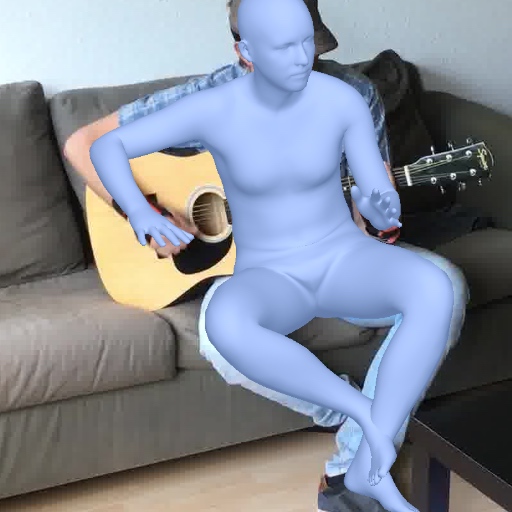}
   \end{subfigure}
    \begin{subfigure}[t]{0.245\textwidth}
    \caption{ \textbf{99$^{\text{th}}$ percentile}}
            \includegraphics[width=0.49\textwidth]{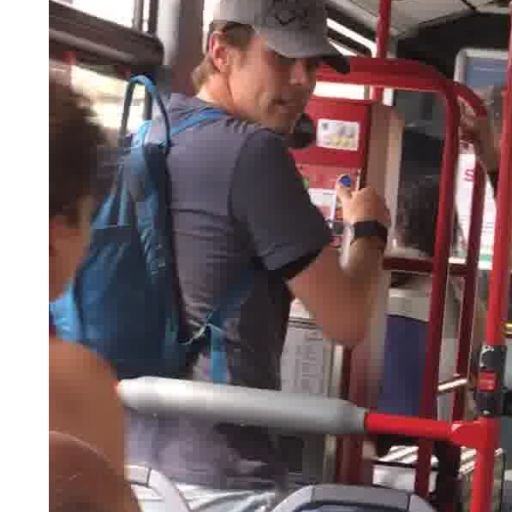}
            \includegraphics[width=0.49\textwidth]{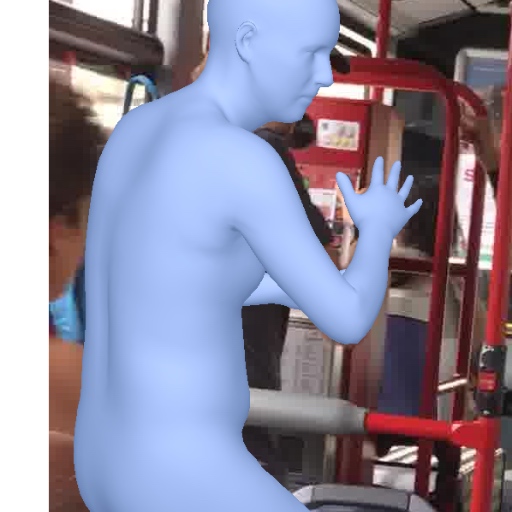}
   \end{subfigure}
   \\
    \begin{subfigure}[t]{0.245\textwidth}
            \includegraphics[width=0.49\textwidth]{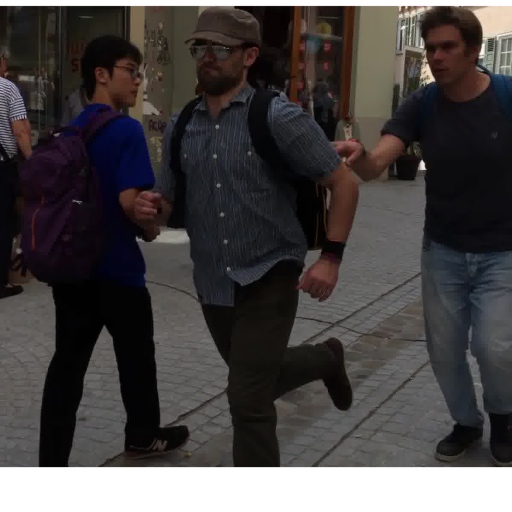}
            \includegraphics[width=0.49\textwidth]{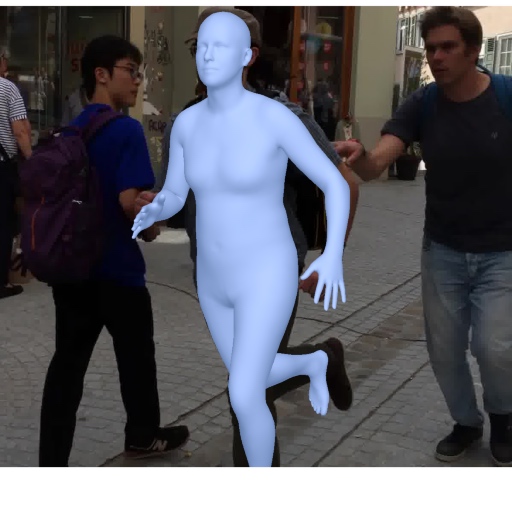}
           \vspace{-6mm}
   \end{subfigure}
    \begin{subfigure}[t]{0.245\textwidth}
            \includegraphics[width=0.49\textwidth]{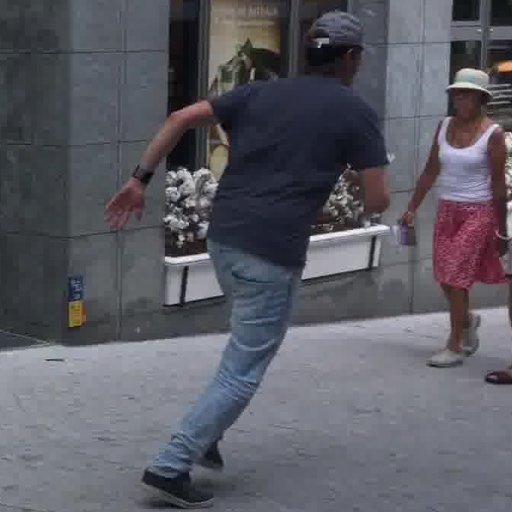}
            \includegraphics[width=0.49\textwidth]{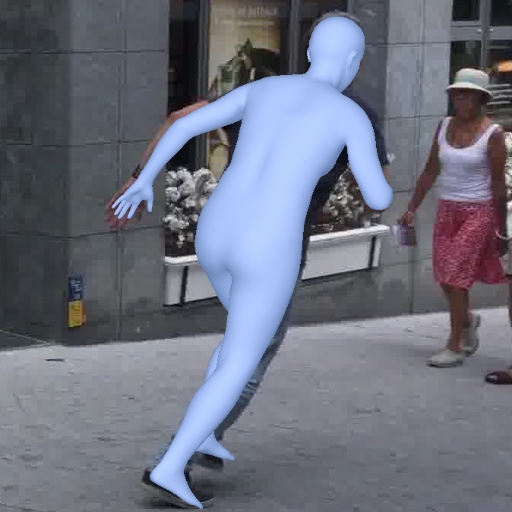}
            \vspace{-6mm}
   \end{subfigure}
    \begin{subfigure}[t]{0.245\textwidth}
            \includegraphics[width=0.49\textwidth]{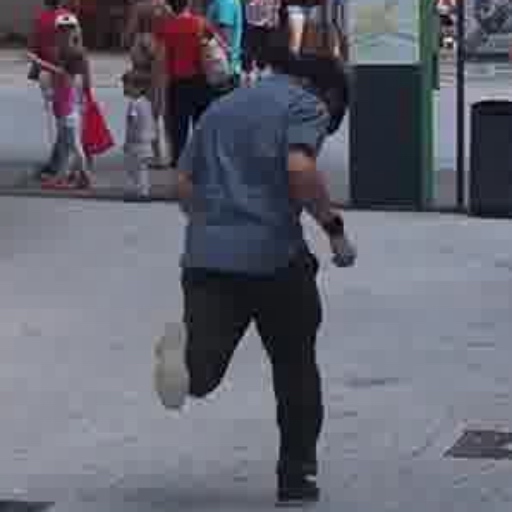}
            \includegraphics[width=0.49\textwidth]{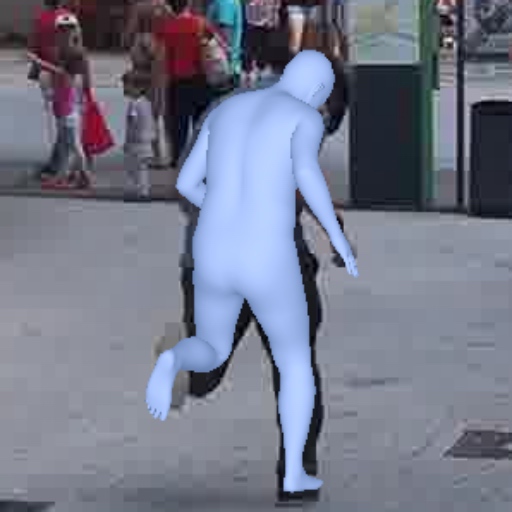}
            \vspace{-6mm}
     \end{subfigure}
    \begin{subfigure}[t]{0.245\textwidth}
            \includegraphics[width=0.49\textwidth]{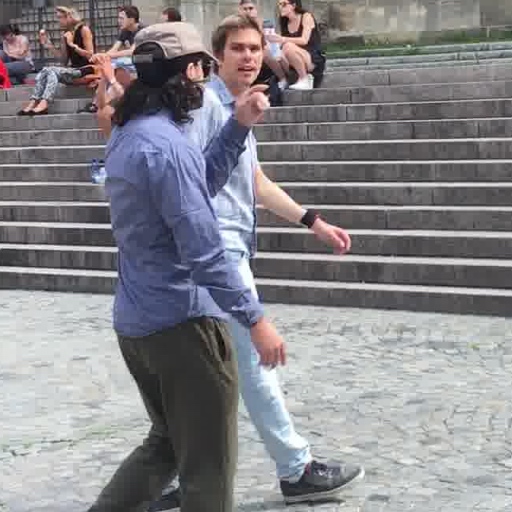}
            \includegraphics[width=0.49\textwidth]{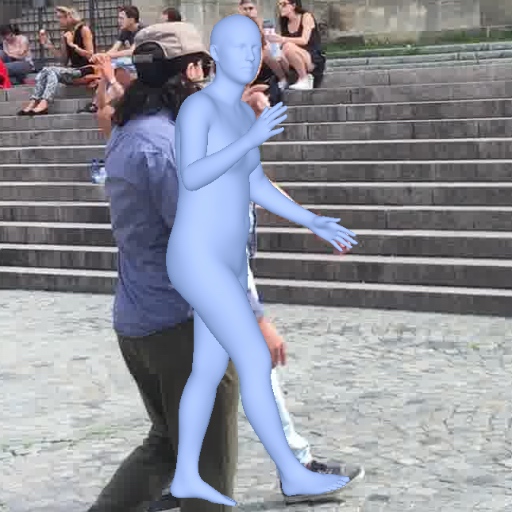}
            \vspace{-6mm}
   \end{subfigure}\\
    \vspace{-2mm}
    \caption{{\bf Qualitative results from ReFit} on 3DPW. Examples are organized by MPJPE percentile. Higher percentile indicates higher error. For example, samples at the 99$^{\text{th}}$ percentile have higher error than 99\% of the examples. The MPJPE at the four percentiles are 50.5mm, 65.4mm, 99.3mm and 158.8mm respectively. Samples at the 99$^{\text{th}}$ percentile often have severe occlusions or cropping.}
    \label{fig:results}
    \vspace{-2mm}
\end{figure*}

We show qualitative results on 3DPW in Figure~\ref{fig:results}. We organize the results by MPJPE percentile, with $X^{\text{th}}$ percentile indicating higher error than $X\%$ of the samples. 

\begin{figure}[ht!]
    \center
    \includegraphics[width=0.242\linewidth]{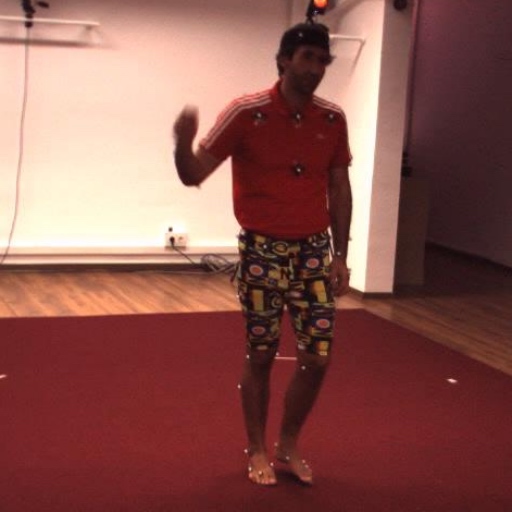} 
    \includegraphics[width=0.242\linewidth]{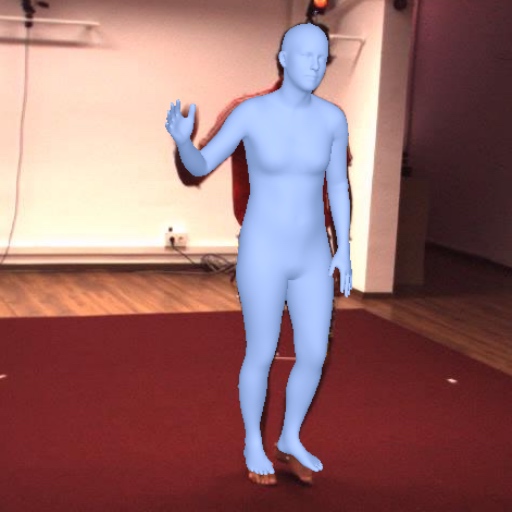} 
    \includegraphics[width=0.242\linewidth]{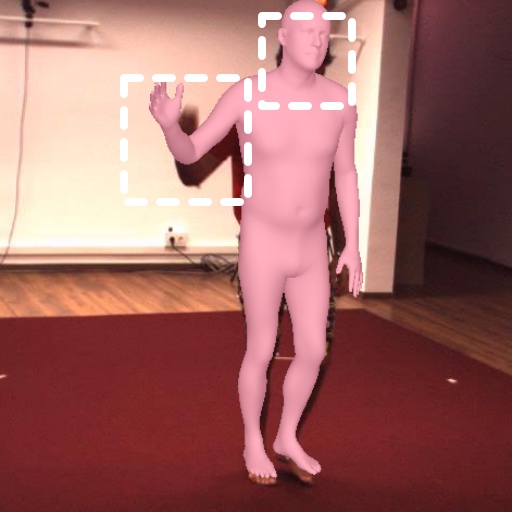} 
    \includegraphics[width=0.242\linewidth]{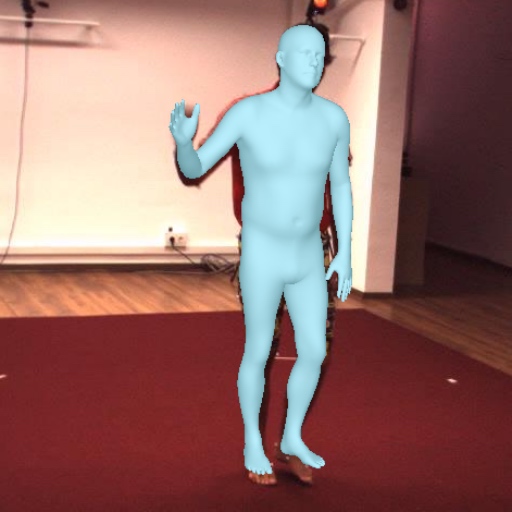} 
     \\
    \includegraphics[width=0.242\linewidth]{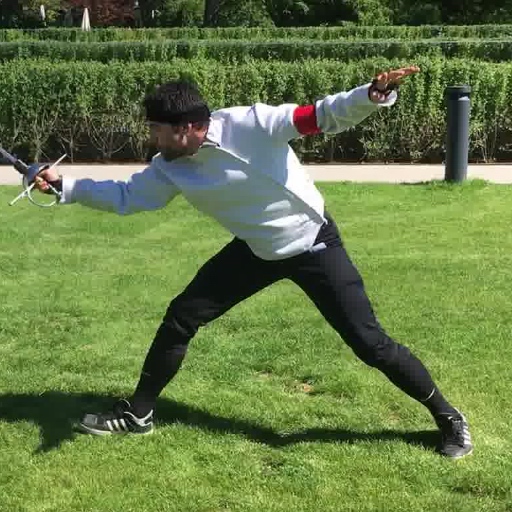} 
    \includegraphics[width=0.242\linewidth]{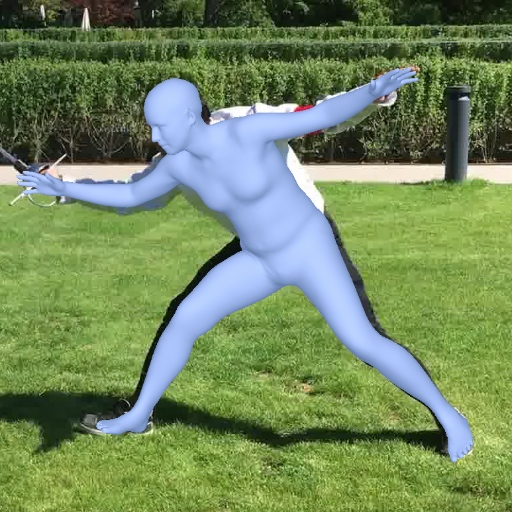} 
    \includegraphics[width=0.242\linewidth]{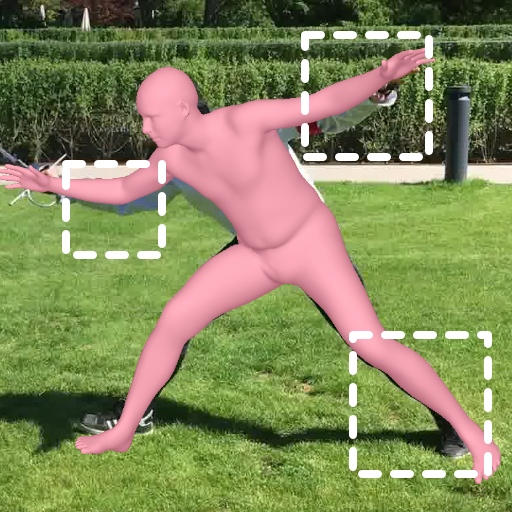} 
    \includegraphics[width=0.242\linewidth]{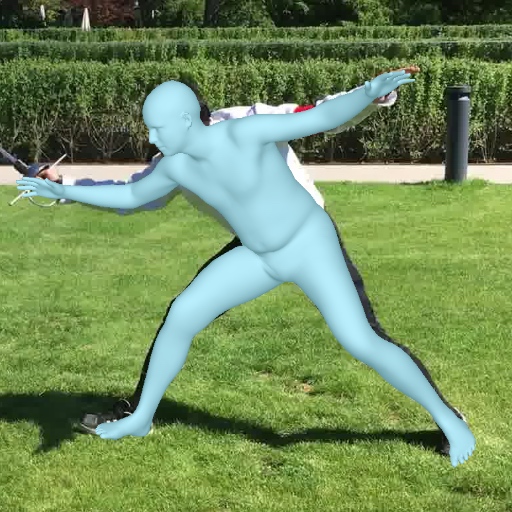} 
\vspace{-5mm}
\caption{\textbf{Shape ReFit.} From left to right are image, ReFit, ReFit with ground truth shape substitute, and Shape ReFit. White boxes highlight misalignment if shape is substituted in place.}
\label{fig:shape}
\vspace{-2mm}
\end{figure}

Overall, we observe accurate reconstructions with good alignment to images throughout different percentiles. We carefully inspect examples at the $99^{\text{th}}$ percentile and find that most examples have severe occlusions. Occlusion by another human produces a second level of complexity. We include more examples in the supplementary. 

\begin{figure}[ht!]
    \center 
    \includegraphics[width=0.242\linewidth]{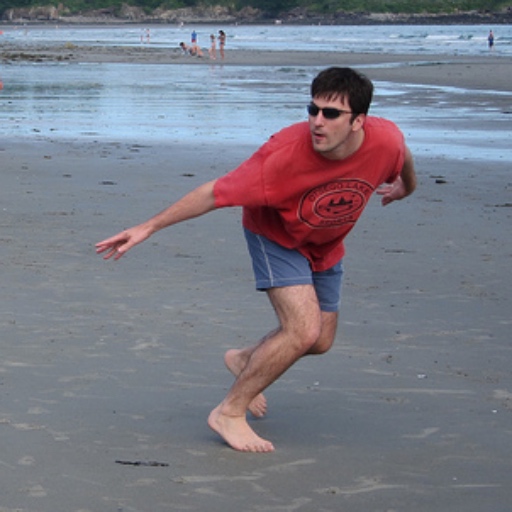} 
    \includegraphics[width=0.242\linewidth]{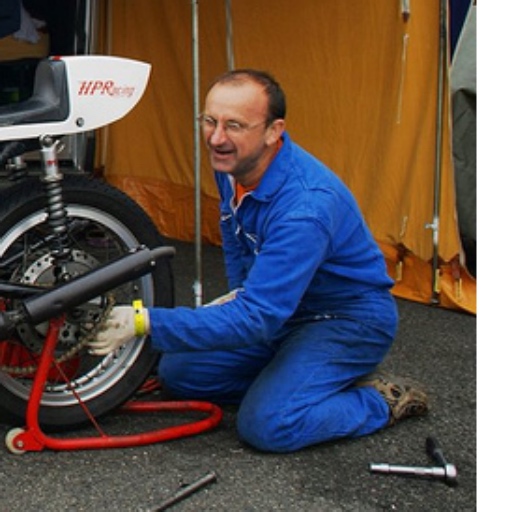} 
    \includegraphics[width=0.242\linewidth]{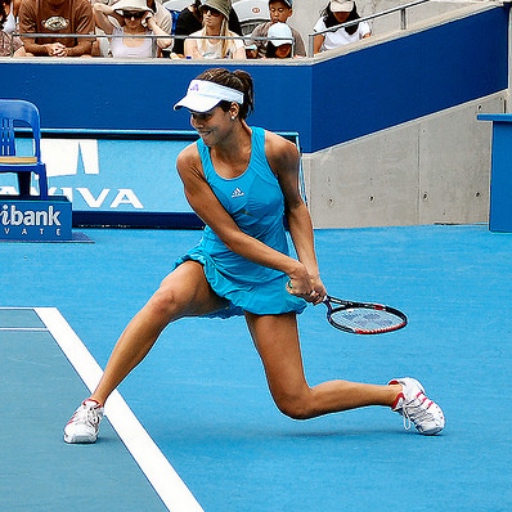} 
    \includegraphics[width=0.242\linewidth]{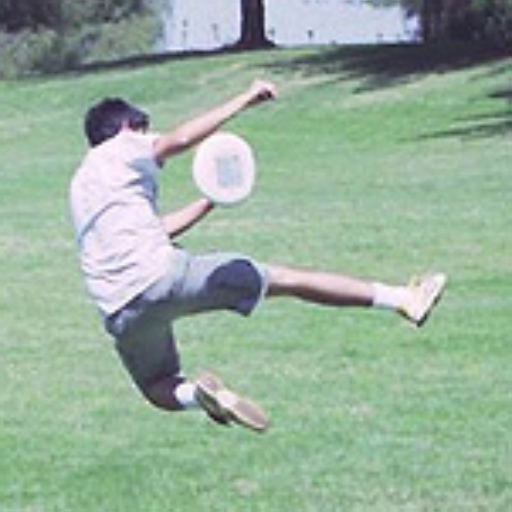} 
     \\
    \includegraphics[width=0.242\linewidth]{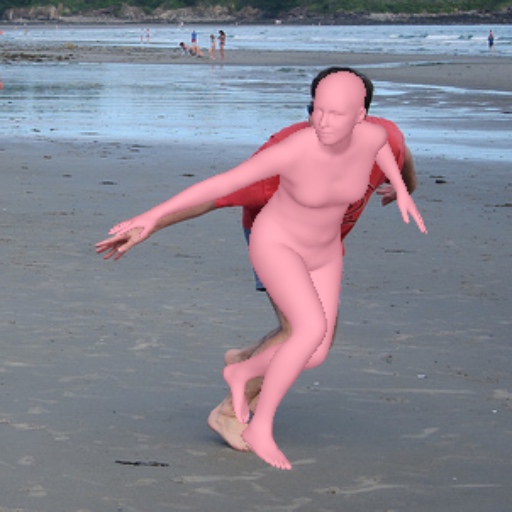} 
    \includegraphics[width=0.242\linewidth]{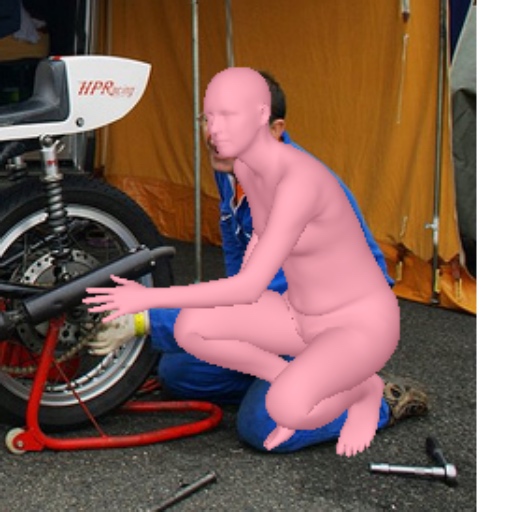} 
    \includegraphics[width=0.242\linewidth]{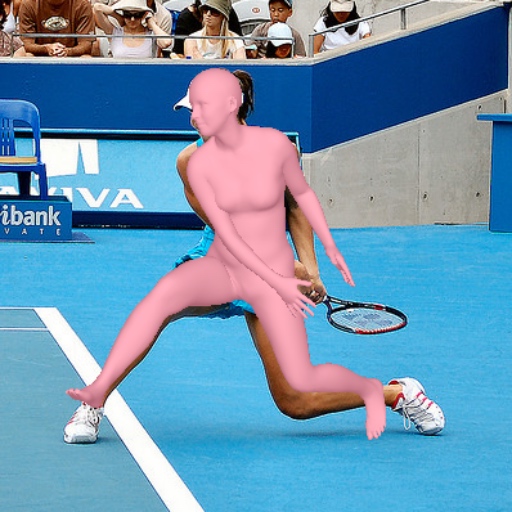} 
    \includegraphics[width=0.242\linewidth]{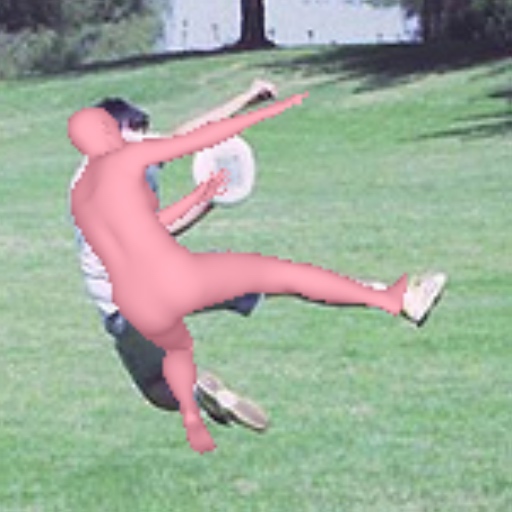} 
     \\
    \includegraphics[width=0.242\linewidth]{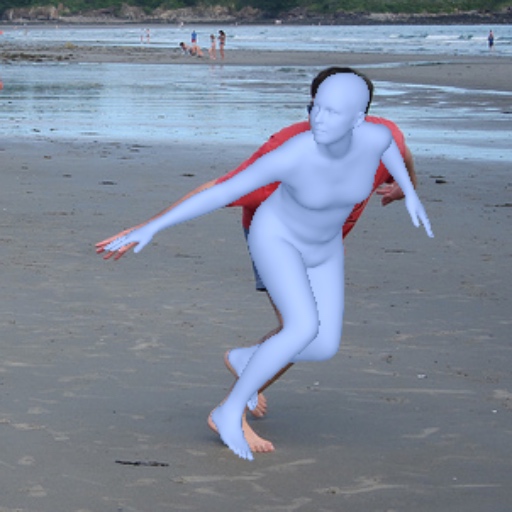} 
    \includegraphics[width=0.242\linewidth]{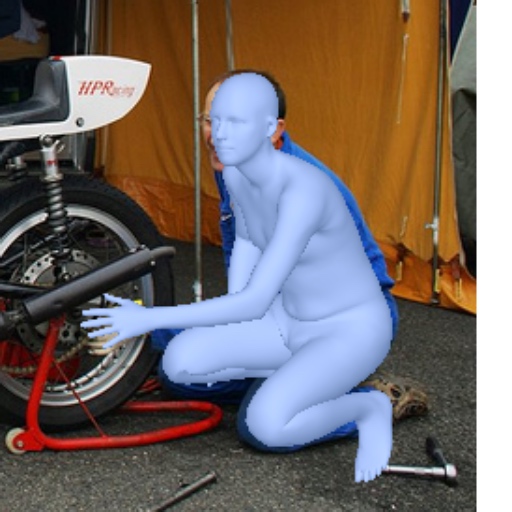} 
    \includegraphics[width=0.242\linewidth]{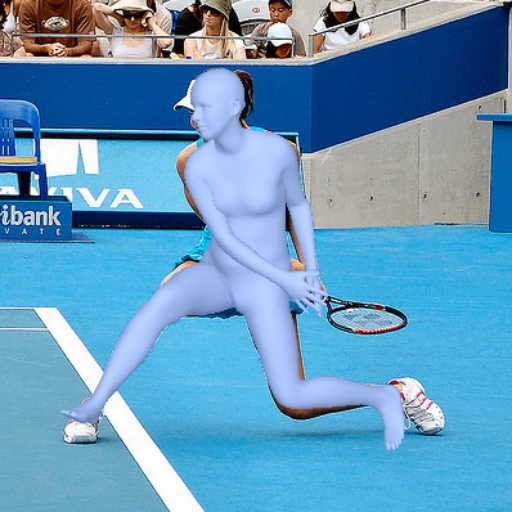} 
    \includegraphics[width=0.242\linewidth]{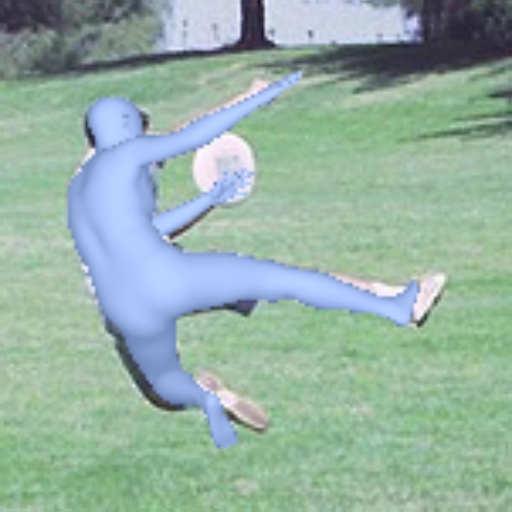} 
    \vspace{-4mm}
\caption{\textbf{Refinement in-the-wild.} Examples from the coco validation set. From top to bottom are images, predictions without refinement update (T=0), and predictions from ReFit (T=5).}
\vspace{-2mm}
\label{fig:coco}
\end{figure}

We provide examples from COCO in Figure~\ref{fig:coco} that highlights the difference between the initial estimation without refinement (T=0) and the final results from ReFit (T=5). 

\subsection{Application Evaluation}
\label{subsec:application_exp}
We evaluate how ReFit performs in the two proposed applications: Shape ReFit and Multi-view ReFit. 

Shape ReFit is more accurate than ReFit, as ground truth shape is used during the fitting process (Tab \ref{tab:shape_fit}). Replacing the final prediction from ReFit with the ground truth shape (ReFit + gt shape) yields similar pose accuracy. However, substituting the shape in place produces misalignment to the image when the predicted shape is noticeably different from the ground truth, as shown in Figure~\ref{fig:shape}. From the qualitative examples, we see that Shape ReFit can indeed fit a pre-acquired shape to its image. 

In the multi-view experiments, we run ReFit and Multi-view ReFit on S9 and S11 from Human3.6M (Tab \ref{tab:multiview}). In the first baseline, we run ReFit on each view separately and report the average reconstruction errors.\begin{table}[ht!]
\centering
\small
\hspace{-3mm}
\tabcolsep=0.85mm

\begin{tabular}{p{0.32\linewidth} cccc@{}} 
\toprule
\multirow{2}{*}{Method} & \multicolumn{2}{c}{3DPW} & \multicolumn{2}{c}{Human3.6M} \\
\cmidrule(lr){2-3} \cmidrule(lr){4-5} 
 & MPJPE & PA-MPJPE & MPJPE & PA-MPJPE \\ 
\midrule
 
ReFit  & 65.8 & 41.0 & 48.4 &  32.2 \\ 
ReFit + gt shape & \textbf{61.7} & \textbf{40.6} & \textbf{41.8} &  \textbf{29.8} \\ 
Shape ReFit & \textbf{61.7} & 40.7 & 44.4 &  32.1  \\ 
    
\bottomrule
\end{tabular}

\vspace{-2mm}
\caption{\small \textbf{Shape ReFit} on 3DPW and Human3.6M. Shape ReFit (Sec.~\ref{subsec:applications}) recovers more accurate poses than ReFit. Substituting shape in place (ReFit + gt shape) has similar accuracy, but produces misalignment (Fig.~\ref{fig:shape}).}
\vspace{-1mm}
\label{tab:shape_fit}
\end{table}\begin{table}[ht!]
\centering
\small
\vspace{-1mm}
\tabcolsep=0.85mm

\begin{tabular}{p{0.42\linewidth} >{\centering\arraybackslash}p{0.18\linewidth} >{\centering\arraybackslash}p{0.20\linewidth} >{\centering\arraybackslash}p{0.15\linewidth}@{}} 
\toprule
 \multirow{2}{*}{Method} & \multicolumn{3}{c}{Human3.6M} \\ \cmidrule(lr){2-4}
 & MPJPE & PA-MPJPE & PVE \\ 
\midrule

ReFit & 52.6 &  34.9 &  66.3 \\ 
ReFit + avg & 41.7 &  29.0 &  54.4 \\ 
Multi-view ReFit (5 iters) & 38.4 &  \textbf{26.6} &  50.0 \\ 
Multi-view ReFit (10 iters) & \textbf{37.5} &  26.9 &  \textbf{48.4} \\ 
\bottomrule
\end{tabular}

\vspace{-2mm}
\caption{\small \textbf{Multi-view ReFit} on Human3.6M. Per-vertex error (PVE) is computed with mesh recovered from MoSH~\cite{loper2014mosh}. Multi-view ReFit uses multi-view information during fitting, and is more accurate than simply averaging the predictions (ReFit + avg).}
\vspace{-3mm}
\label{tab:multiview}
\end{table}In the second baseline, we run ReFit on each view separately but average the final predictions. Averaging the predictions from multiple views improves the results, particularly in MPJPE, due to a better global rotation estimation. The proposed multi-view ReFit produces the best results. The improvement is more evident in terms of the per-vertex error (PVE), where we see a 27\% improvement over the baseline. In this setting, we also see the benefit of running for more update iterations, with 10 iterations slightly better than 5 iterations.

\begin{figure*}[!t]
\captionsetup[subfigure]{labelformat=empty}
    \centering
    \vspace{-1mm}
    \begin{subfigure}[t]{0.245\textwidth}
    \caption{\textbf{Two Views} }
    \includegraphics[width=0.49\textwidth]{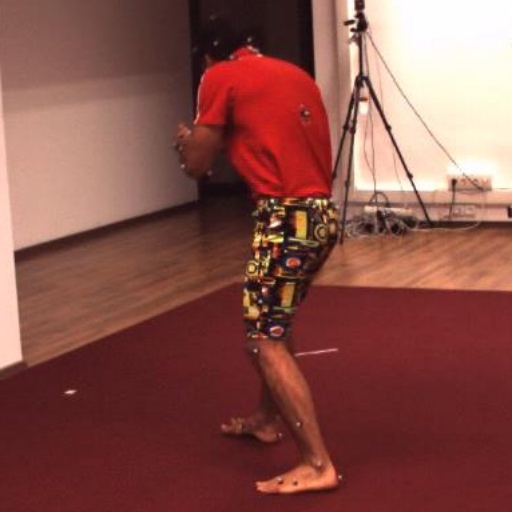}
    \includegraphics[width=0.49\textwidth]{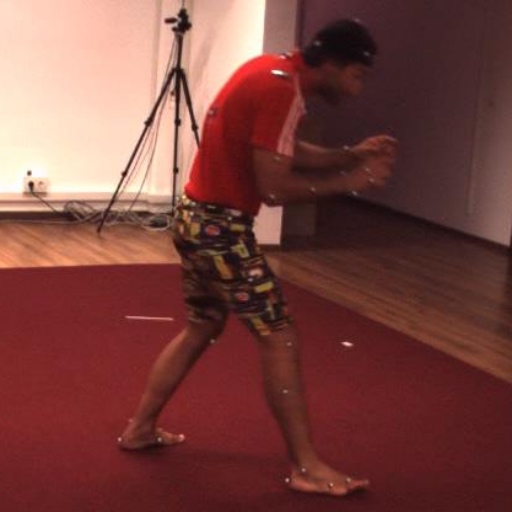}
    \end{subfigure}
    \begin{subfigure}[t]{0.245\textwidth}
    \caption{ \textbf{Multi-view ReFit}}
    \includegraphics[width=0.49\textwidth]{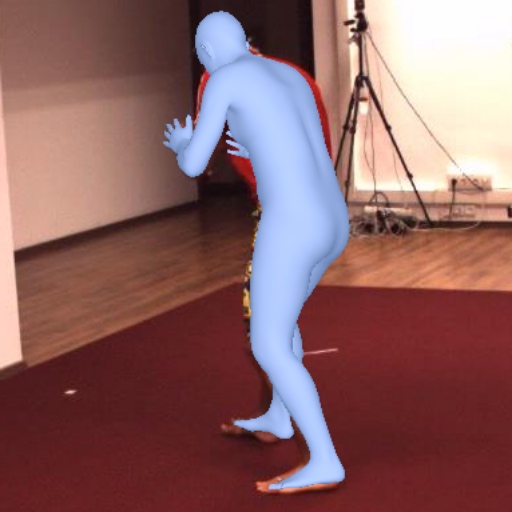}
    \includegraphics[width=0.49\textwidth]{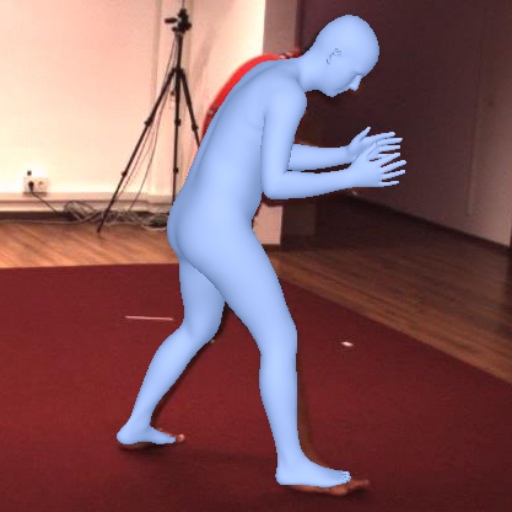}
    \end{subfigure}
    \begin{subfigure}[t]{0.1225\textwidth}
    \caption{}
    \includegraphics[width=0.98\textwidth]{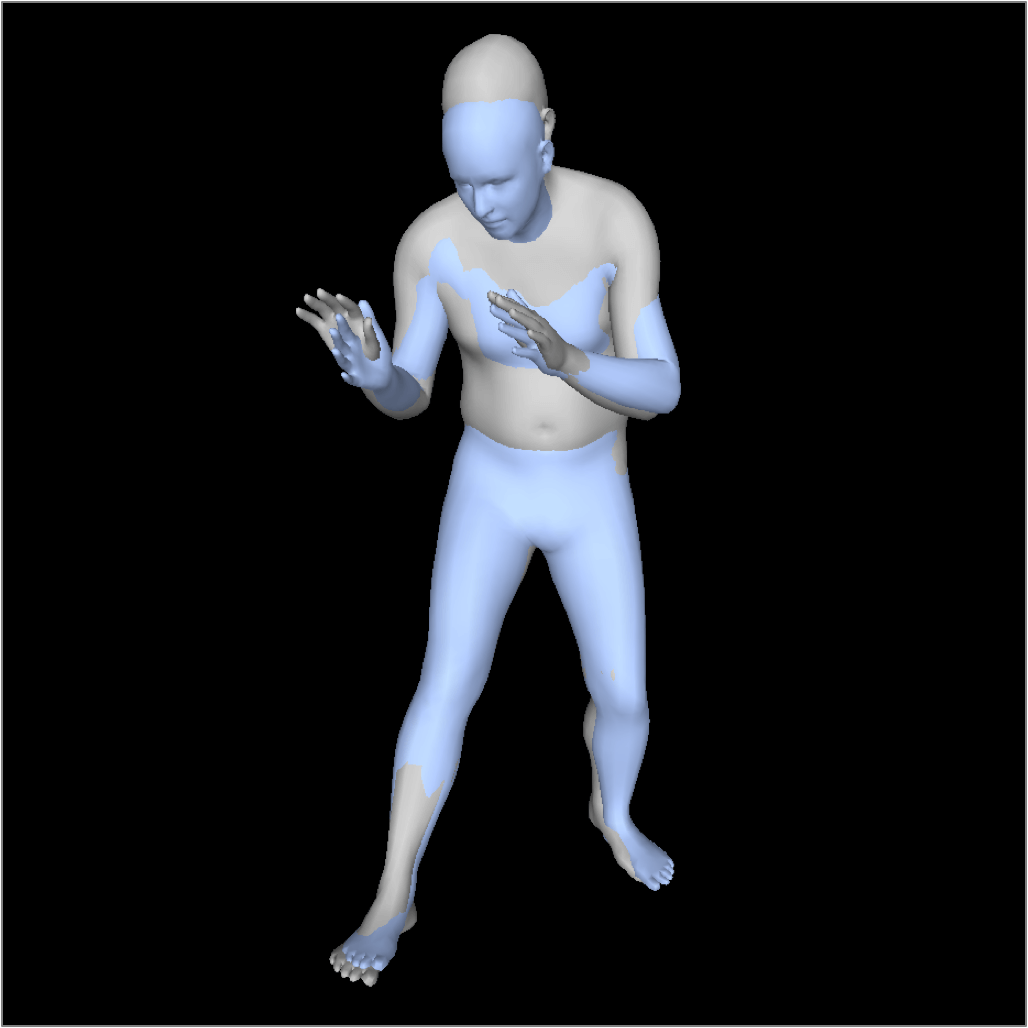}
    \end{subfigure}
    \begin{subfigure}[t]{0.245\textwidth}
    \caption{\textbf{Multi-view + Shape ReFit}}
    \includegraphics[width=0.49\textwidth]{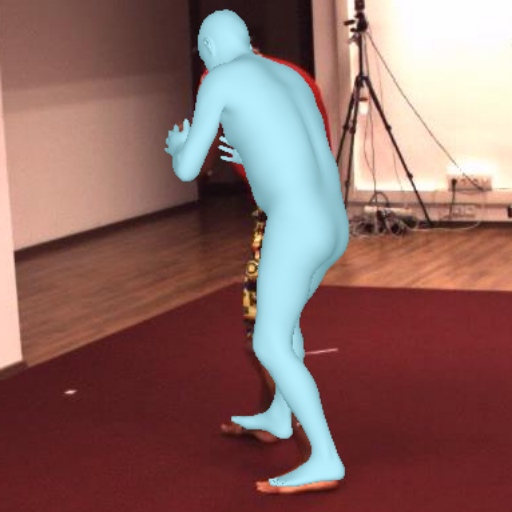}
    \includegraphics[width=0.49\textwidth]{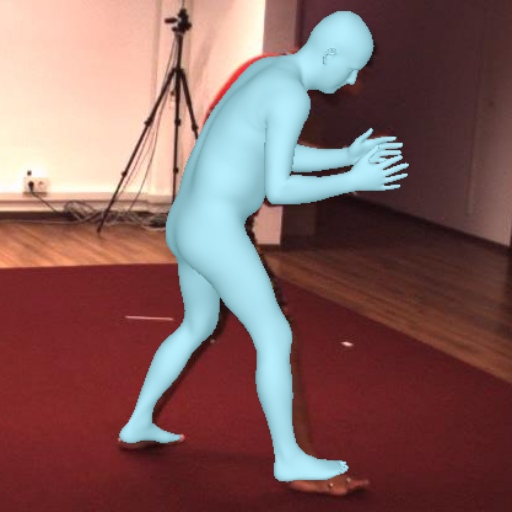}
    \end{subfigure}
    \begin{subfigure}[t]{0.1225\textwidth}
    \caption{}
    \includegraphics[width=0.98\textwidth]{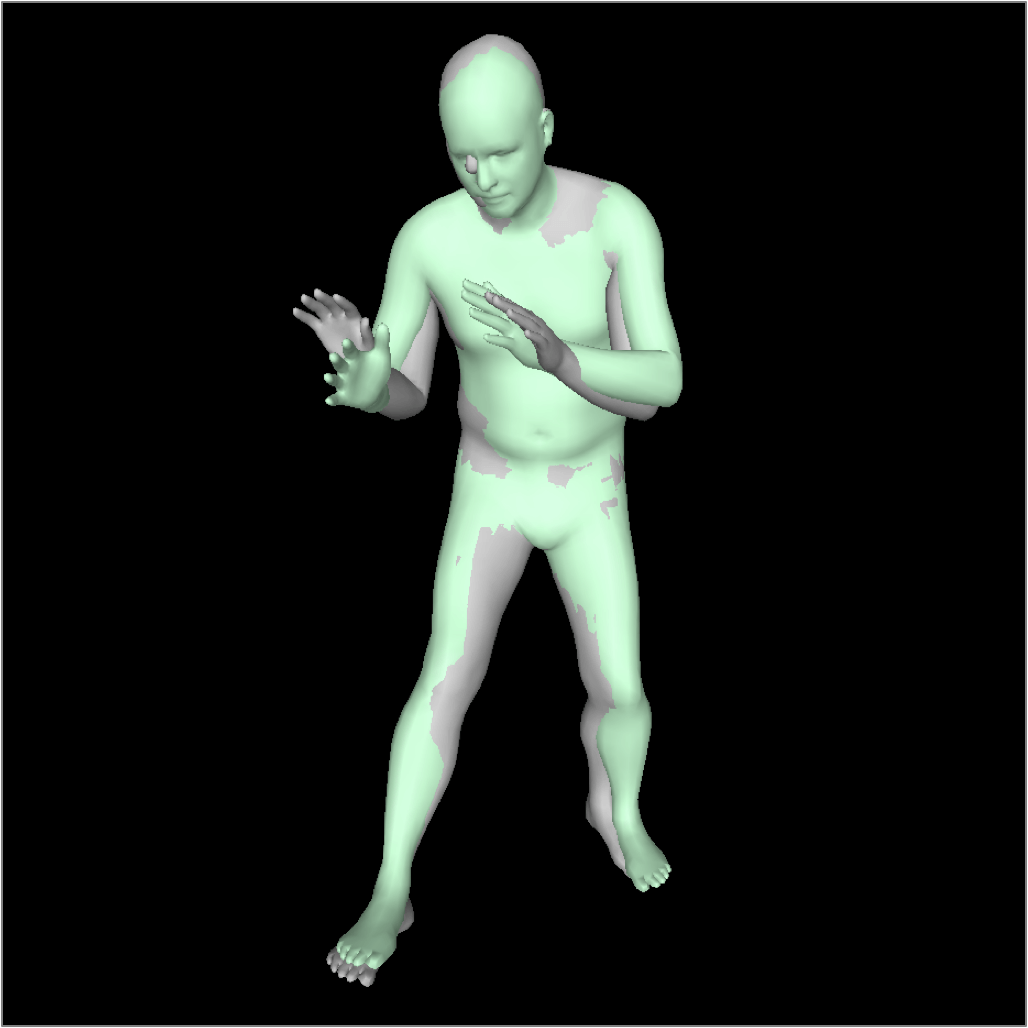}
    \end{subfigure}
    \\
    \begin{subfigure}[t]{0.245\textwidth}
    \includegraphics[width=0.49\textwidth]{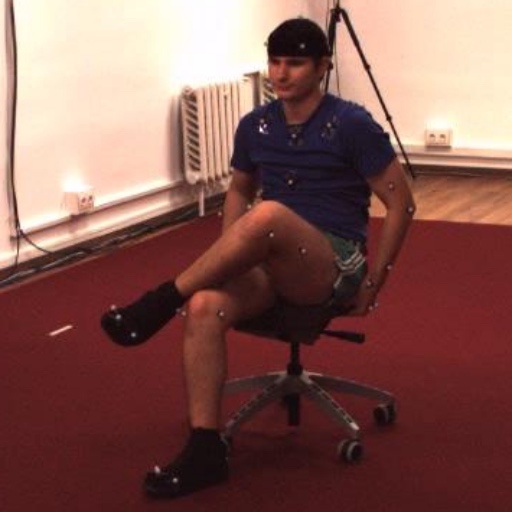}
    \includegraphics[width=0.49\textwidth]{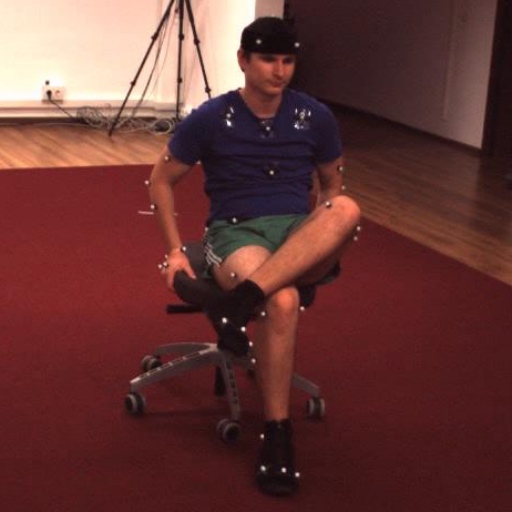}
    \vspace{-6mm}
    \end{subfigure}
    \begin{subfigure}[t]{0.245\textwidth}
    \includegraphics[width=0.49\textwidth]{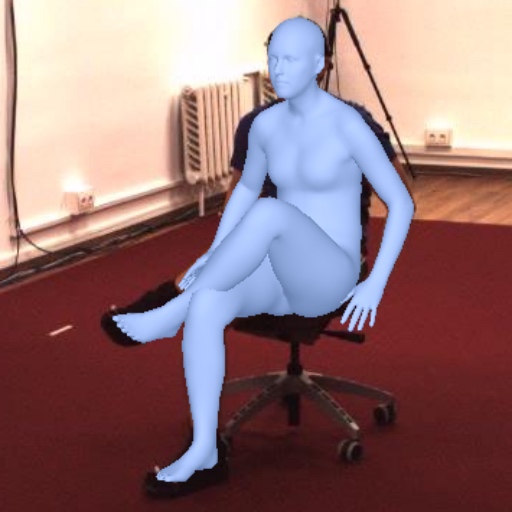}
    \includegraphics[width=0.49\textwidth]{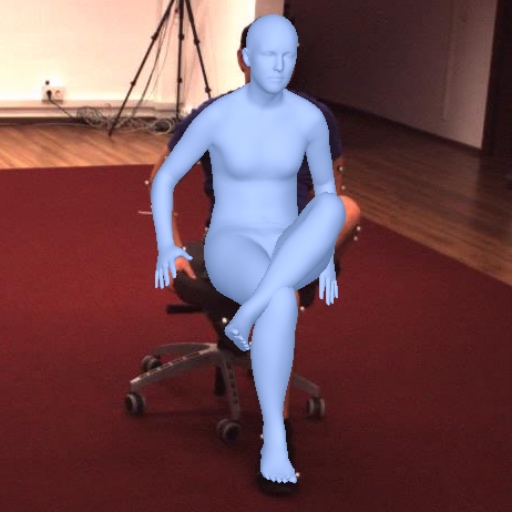}
    \vspace{-6mm}
    \end{subfigure}
    \begin{subfigure}[t]{0.1225\textwidth}
    \includegraphics[width=0.98\textwidth]{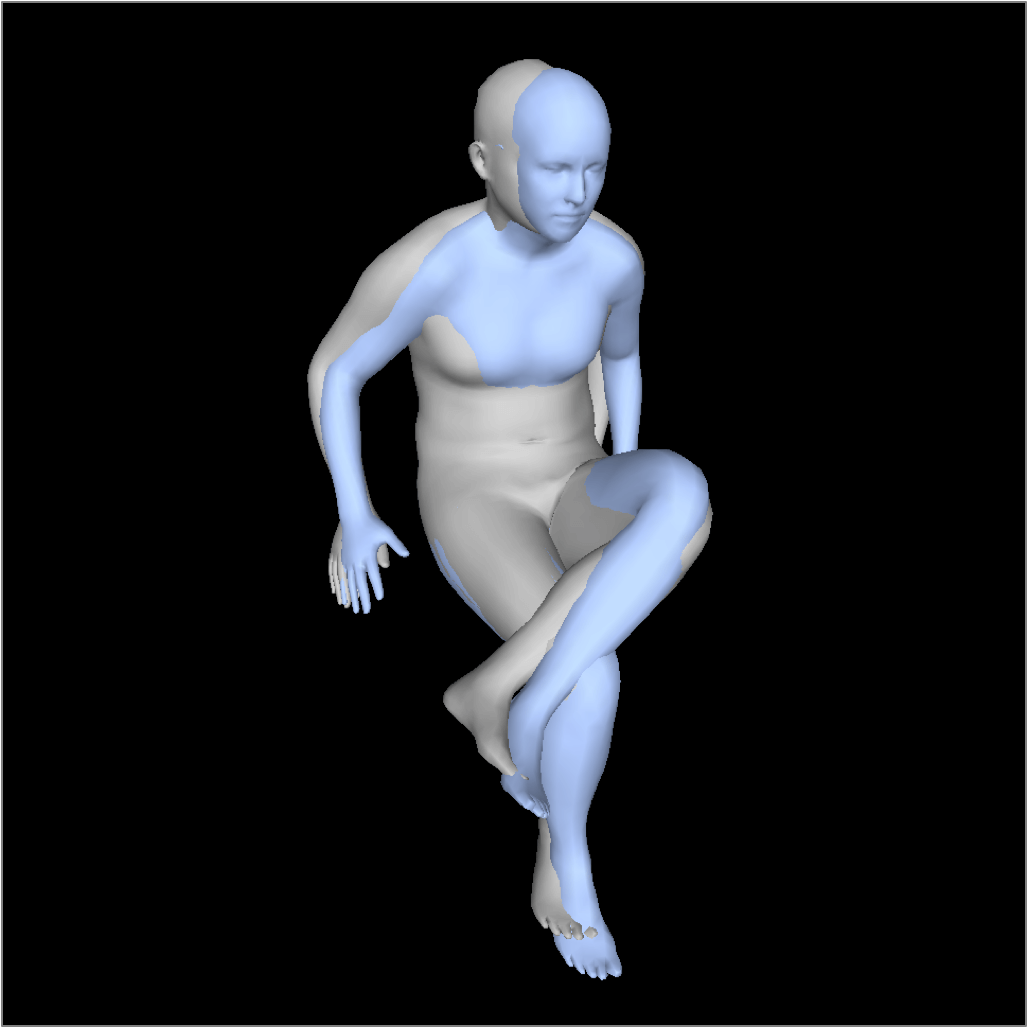}
    \vspace{-6mm}
    \end{subfigure}
    \begin{subfigure}[t]{0.245\textwidth}
    \includegraphics[width=0.49\textwidth]{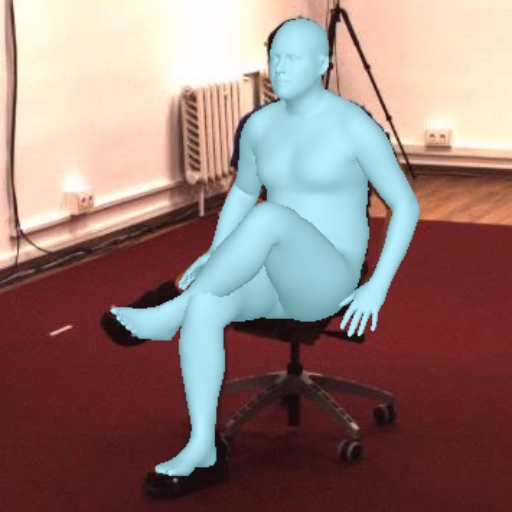}
    \includegraphics[width=0.49\textwidth]{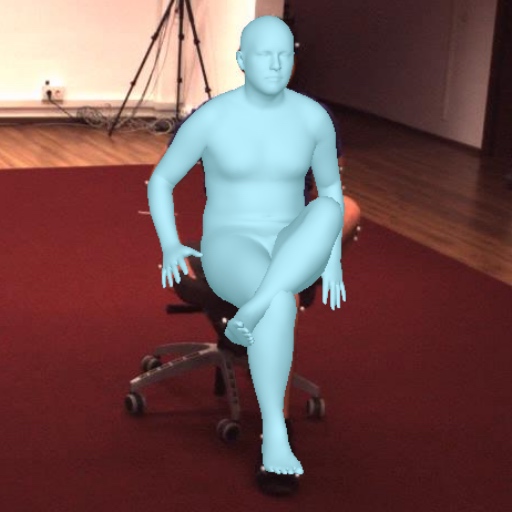}
    \vspace{-6mm}
    \end{subfigure}
    \begin{subfigure}[t]{0.1225\textwidth}
    \includegraphics[width=0.98\textwidth]{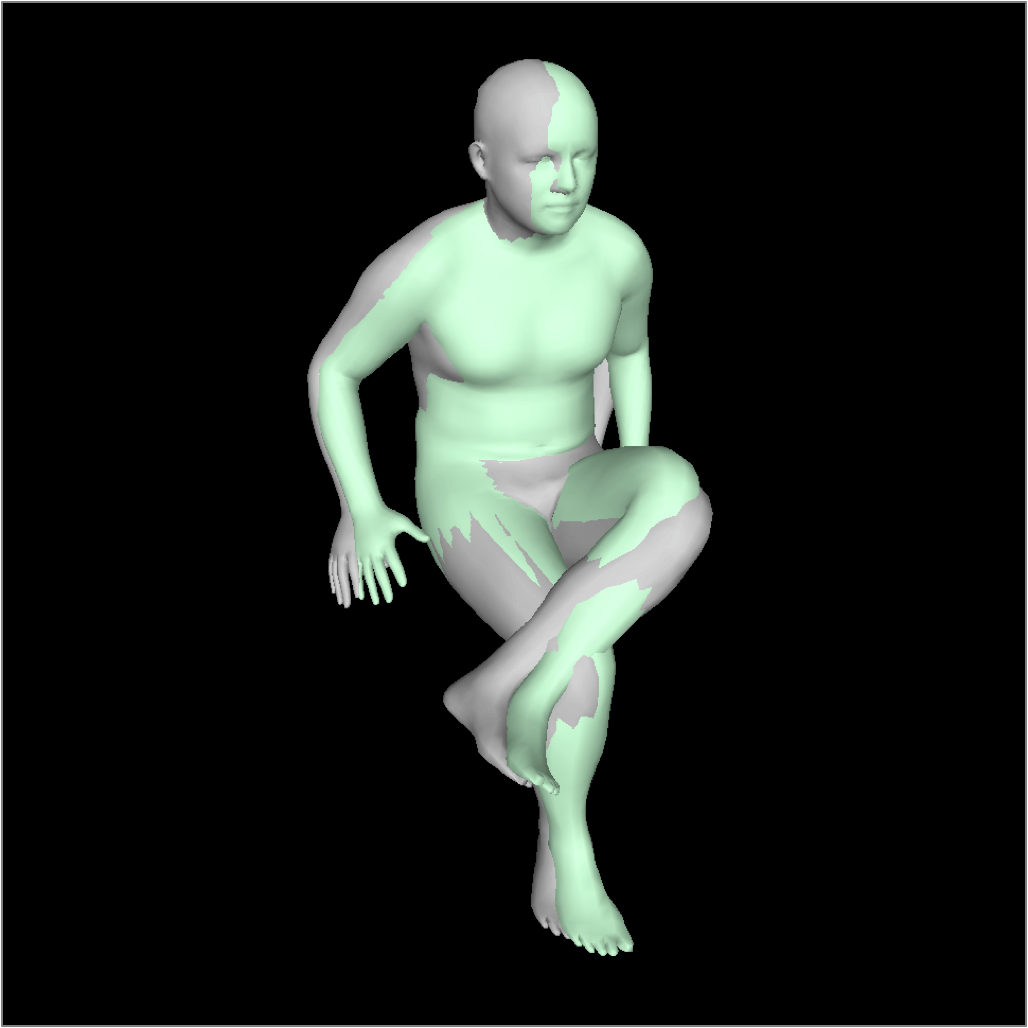}
    \vspace{-6mm}
    \end{subfigure}\\
    \vspace{-2mm}
    \caption{{\bf Multi-vew ReFit.} We show results for 2 of 4 views. In the middle (blue meshes), multi-view ReFit reconstructs accurate poses compare to the ground truth grey meshes from MoSH~\cite{loper2014mosh}. On the right (green meshes), assuming ground truth shapes are pre-fitted and available, multi-view ReFit produces results that are very close to MoSH.}
    \label{fig:multiview}
    \vspace{-2mm}
\end{figure*}

\begin{figure}[ht!]
    \center 
    \includegraphics[width=0.242\linewidth]{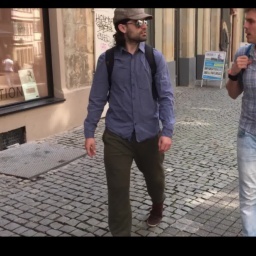} 
    \includegraphics[width=0.242\linewidth]{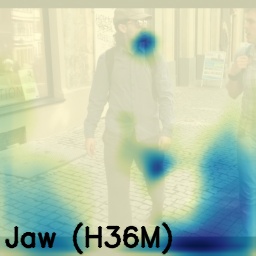} 
    \includegraphics[width=0.242\linewidth]{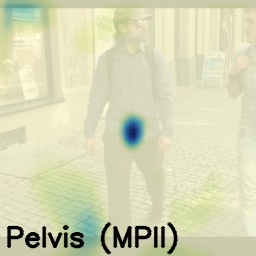} 
    \includegraphics[width=0.242\linewidth]{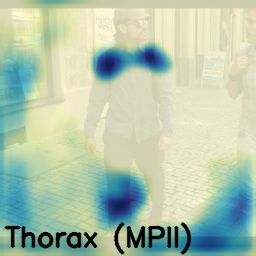} 
     \\
    \includegraphics[width=0.242\linewidth]{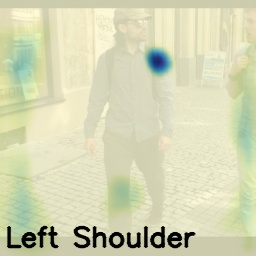} 
    \includegraphics[width=0.242\linewidth]{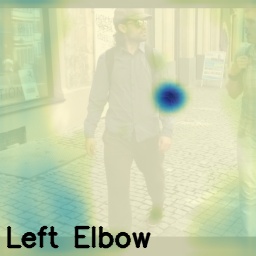} 
    \includegraphics[width=0.242\linewidth]{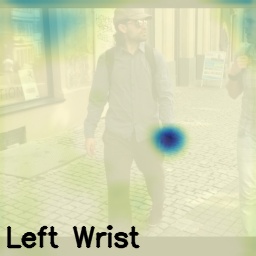} 
    \includegraphics[width=0.242\linewidth]{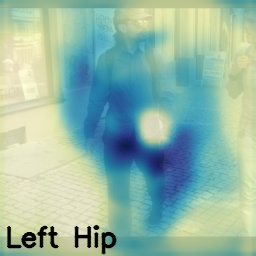} 
     \\
    \includegraphics[width=0.242\linewidth]{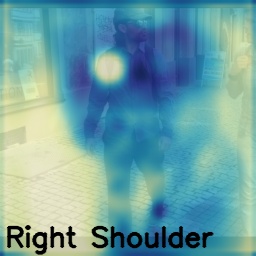} 
    \includegraphics[width=0.242\linewidth]{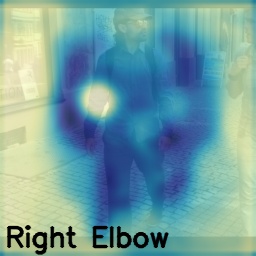} 
    \includegraphics[width=0.242\linewidth]{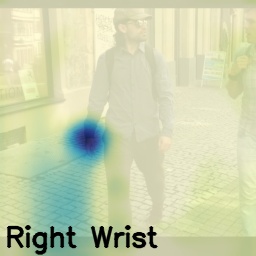} 
    \includegraphics[width=0.242\linewidth]{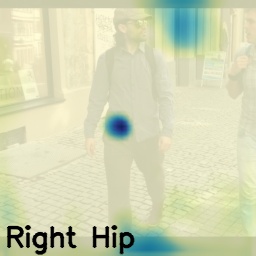} 
    \vspace{-4mm}
\caption{\textbf{Per-keypoint feature map.} Each channel corresponds to the learned features of a keypoint. Here we show feature maps from the semantic keypoint model. The network learns a peak (blue) or a valley (yellow surrounded by blue) response around the keypoint, without direct supervision. 
}
\vspace{-2mm}
\label{fig:heatmap}
\end{figure}

We show qualitative results for Multi-view ReFit in Figure~\ref{fig:multiview}. We compare the results against the mesh recovered using MoSH~\cite{loper2014mosh}. Multi-view ReFit produces accurate 3D poses. We further test combining Multi-view with Shape ReFit. The recovered mesh is very close to MoSH results. MoSH is a proprietary optimization procedure that uses multi-view video and localized 3D markers, while our method only assumes multi-view images and optionally a pre-fitted shape. This result points to an alternative mocap procedure, where one can pre-fit the shape to an A-posed subject with optimization and use Multi-view + Shape ReFit for motion capture. The result can be used on its own or as initialization for optimization. A detailed comparison with MoSH is left for future work. 

\subsection{Visualization of Feature Maps}
\label{subsec:feature_map}
 Feature maps from the proposed model are slightly blurry and harder to visualize, as they have positive and negative values. We train an alternative model, where we add a ReLU operator on the feature maps. This modification decreases the performance slightly, as the feature maps become less expressive (strictly positive), but makes them easier to visualize. Figure~\ref{fig:heatmap} shows examples from the model with ReLU, and we include more examples from both models in the supplementary. 

We show feature maps from the semantic keypoint model, as semantic keypoints are more interpretable than markers. The model learns meaningful features around the corresponding keypoints, often as peaks or valleys. When a keypoint is reprojected onto the feature map, the peak or valley provides directional information that helps the next network layer infer where the keypoint should move on the 2D plane. This signal, combined with signals from other keypoints, can then be converted to updates for the 3D pose.

\section{Conclusion}
\label{sec:conclusion}
We have presented ReFit, the Recurrent Fitting Network that iteratively fits the parametric human model to images to recover the pose and shape of people. ReFit mirrors traditional model fitting, but learns the objective and update rules from data end-to-end. ReFit utilizes parallel GRU units that disentangle the learned update rules. Moreover, we demonstrate ReFit as a learned optimizer for multi-view fitting and shape fitting. ReFit is efficient to train and achieves state-of-the-art accuracy in challenging datasets. 

\smallbreak
\begin{normalsize}
\noindent\textbf{Acknowledgements:} We gratefully acknowledge support by the NSF grants FRR-2220868 and IIS-RI-2212433.
\end{normalsize}

{\small
\bibliographystyle{ieee_fullname}
\bibliography{11_references}
}

\end{document}